\documentclass[sigconf]{acmart}

\usepackage{graphicx}
\usepackage{multirow}
\usepackage{xspace}
\usepackage{threeparttable}
\usepackage{enumitem}
\usepackage{makecell}
\usepackage{bbding}
\usepackage{caption}
\usepackage{subfig}
\usepackage{booktabs}
\usepackage[export]{adjustbox}

\usepackage{amsmath,amsfonts,bm}

\def\eqref#1{equation~\ref{#1}}

\def\1{\bm{1}}

\def\vh{{\bm{h}}}

\def\vp{{\bm{p}}}

\def\vx{{\bm{x}}}

\def\vz{{\bm{z}}}

\def\mA{{\bm{A}}}

\def\mH{{\bm{H}}}

\def\mX{{\bm{X}}}

\def\mZ{{\bm{Z}}}

\DeclareMathAlphabet{\mathsfit}{\encodingdefault}{\sfdefault}{m}{sl}
\SetMathAlphabet{\mathsfit}{bold}{\encodingdefault}{\sfdefault}{bx}{n}

\newcommand{\vol}{\operatorname{vol}}

\newcommand{\vpara}[1]{\vspace{0.04in}\noindent\textbf{#1}\xspace}

\newcommand{\hide}[1]{} 

\newcommand{\model}{GraphMAE2\xspace}

\AtBeginDocument{%
  \providecommand\BibTeX{{%
    \normalfont B\kern-0.5em{\scshape i\kern-0.25em b}\kern-0.8em\TeX}}}

\setcopyright{acmcopyright}
\hide{
\copyrightyear{2023}
\acmYear{2023}
\acmDOI{10.1145/3543507.3583379}
\acmConference[WWW'23]{Make sure to enter the correct
  conference title from your rights confirmation emai}{April 30--May 4, 2023}{Austin, Texas, USA}
\acmPrice{15.00}
\acmISBN{978-1-4503-XXXX-X/18/06}
}

\copyrightyear{2023}
\acmYear{2023}
\setcopyright{rightsretained}
\acmConference[WWW '23]{Proceedings of the ACM Web Conference 2023}{May 1--5, 2023}{Austin, TX, USA}
\acmBooktitle{Proceedings of the ACM Web Conference 2023 (WWW '23), May 1--5, 2023, Austin, TX, USA}
\acmDOI{10.1145/3543507.3583379}
\acmISBN{978-1-4503-9416-1/23/04}

\begin{document}

\title{\model: A Decoding-enhanced Masked Self-supervised Graph Learner}
\title{\model: A Decoding-Enhanced Masked Self-Supervised Graph Learner}



\author{Zhenyu Hou}
\affiliation{Tsinghua University\country{China}}
\email{houzy21@mails.tsinghua.edu.cn}

\author{Yufei He}
\affiliation{Beijing Institute of Technology\country{China}}
\email{yufei.he@bit.edu.cn}
\authornote{
This work was done when the author was visiting Tsinghua University. 
}

\author{Yukuo Cen}
\affiliation{Tsinghua University\country{China}}
\email{cyk20@mails.tsinghua.edu.cn}

\author{Xiao Liu}
\affiliation{Tsinghua University\country{China}}
\email{liuxiao21@mails.tsinghua.edu.cn}

\author{Yuxiao Dong}
\authornote{Yuxiao Dong and Jie Tang are the corresponding authors.}
\affiliation{Tsinghua University\country{China}}
\email{yuxiaod@tsinghua.edu.cn}

\author{Evgeny Kharlamov}
\affiliation{Bosch Center for Artificial Intelligence\country{Germany}}
\email{evgeny.kharlamov@de.bosch.com}

\author{Jie Tang}
\authornotemark[2]
\affiliation{Tsinghua University\country{China}}
\email{jietang@tsinghua.edu.cn}

 \renewcommand{\authors}{Zhenyu Hou, Yufei He, Yukuo Cen, Xiao Liu, Yuxiao Dong, Evgeny Kharlamov, Jie Tang}

\renewcommand{\shortauthors}{Zhenyu Hou, Yufei He, Yukuo Cen, Xiao Liu, Yuxiao Dong, Evgeny Kharlamov, Jie Tang}

\begin{abstract}

Graph self-supervised learning (SSL), including contrastive and generative approaches, offers great potential to address the fundamental challenge of label scarcity in real-world graph data. 
Among both sets of graph SSL techniques, the masked graph autoencoders (e.g., GraphMAE)---one type of generative methods---have recently produced promising results. 
The idea behind this is to reconstruct the node features (or structures)---that are randomly masked from the input---with the autoencoder architecture. 
However, the performance of masked feature reconstruction naturally relies on the discriminability of the input features and is usually vulnerable to disturbance in the features. 
In this paper, we present a masked self-supervised learning framework\footnote{The code is available at:
\url{https://github.com/THUDM/GraphMAE2}.} \model with the goal of overcoming this issue. 
The idea is to impose regularization on feature reconstruction for graph SSL.  
Specifically, we design the strategies of multi-view random re-mask decoding and latent representation prediction to regularize the feature reconstruction. 
The multi-view random re-mask decoding is to introduce randomness into reconstruction in the feature space, while the latent representation prediction is to enforce the reconstruction in the embedding space. 
Extensive experiments show that \model can consistently generate top results on various public datasets, including at least 2.45\% improvements over state-of-the-art baselines on ogbn-Papers100M with 111M nodes and 1.6B edges. 
\end{abstract}

\ccsdesc[500]{Computing methodologies~Learning latent representations}
\ccsdesc[500]{Information systems~Data mining}

\keywords{Graph Neural Networks; Self-Supervised Learning; Graph Representation Learning; Pre-Training}

\maketitle

\section{Introduction}
Graph neural networks (GNNs) have found widespread adoption in learning representations for graph-structured data. 
The success of GNNs has thus far mostly occurred in (semi-) supervised settings, in which task-specific labels are used as the supervision information, such as GCN~\cite{thomas2017gcn}, GAT~\cite{petar2018gat}, and GraphSAGE~\cite{hamilton2017inductive}. 
However, it is often arduously difficult to obtain sufficient labels in real-world scenarios, especially for billion-scale graphs~\cite{hu2020gpt,hu2019strategies}.

\begin{figure}
    \centering
    \includegraphics[width=0.5\textwidth]{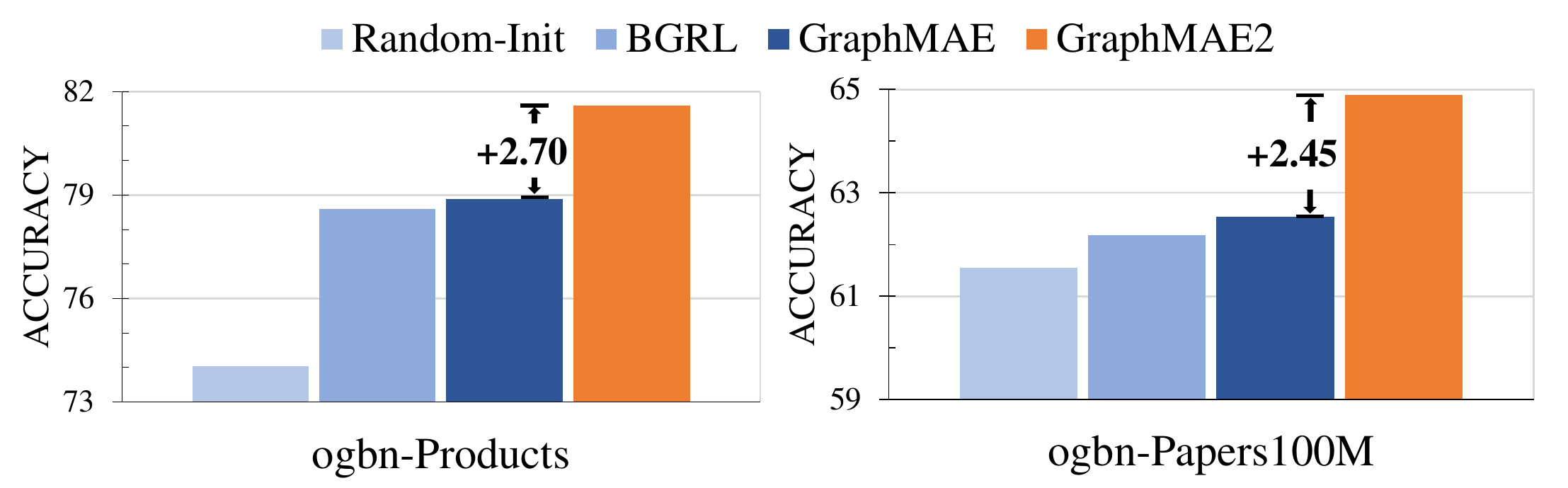}
    \caption{Linear probing results on ogbn-Products and ogbn-Papers100M. \textmd{\model achieves a significant advantage over previous graph SSL methods on benchmarks with millions of nodes. }}
    \label{fig:head_fg}
\end{figure}

One natural solution to this challenge is to perform self-supervised learning (SSL) on graphs~\cite{liu2021self}, where graph models (e.g., GNNs) are supervised by labels that are automatically constructed from the input graph data. 
Along this line, generative SSL models that aim to generate one part of the input graph from another part have received extensive exploration~\cite{kipf2016variational,wang2017mgae,pan2018adversarially,hu2020gpt,cui2020adaptive}. 
Straightforwardly, it first corrupts the input graph by masking node features or edges and then learns to recover the original input.

Under the masked prediction framework, a very recent work introduces a masked graph autoencoder GraphMAE~\cite{hou2022graphmae} for generative SSL on graphs, which yields outperformance over various baselines on 21 datasets for different tasks. 
Generally, an autoencoder is made up of an encoder, code/embeddings, and a decoder. 
The encoder maps the input to embeddings, and the decoder aims to reconstruct the input based on the embeddings under a reconstruction criterion. 
The main idea of GraphMAE is to reconstruct the input node features that are randomly masked before encoding by using an autoencoding architecture.
Its technical contribution lies in the design of 1) masked feature reconstruction and 2) fixed re-mask decoding, wherein the encoded embeddings of previously-masked nodes are masked again before feeding into the decoder.

Despite GraphMAE's promising performance, the reconstruction of masked features fundamentally relies on the discriminability~\cite{chien2021node,wei2022masked} of the input node features, i.e., the extent to which the node features are distinguishable. 
In practice, the features of nodes in a graph are usually generated from data that is associated with each node, such as the embeddings of content posted by users in a social network, making them an approximate description of nodes and thus less discriminative. 
Note that in vision or language studies, the reconstruction targets are usually a natural description of the data, i.e., pixels of an image and words of a document. 
Table ~\ref{tab:pre_exp} further shows that the performance of GraphMAE drops more significantly than the supervised counterpart when using less discriminative node features (w/ PCA). 
In other words, GraphMAE, as a generative SSL framework with feature reconstruction, is relatively more vulnerable to the disturbance of features.

In this work, we present \model with the goal of improving  feature reconstruction for graph SSL. 
The idea is to impose regularization on target reconstruction.  
To achieve this, we introduce two decoding strategies: \textit{multi-view random re-mask decoding} for reducing the overfitting to the input features, and \textit{latent representation prediction} for having more informative targets. 

First, instead of fixed re-mask decoding used in GraphMAE---re-masking the encoded embeddings of masked nodes, we propose to introduce randomness into input feature reconstruction with \textit{multi-view random re-mask decoding}. 
That is, the encoded embeddings are randomly re-masked multiple times, and their decoding results are all enforced to recover input features. 
Second, we propose \textit{latent representation prediction}, which attempts to reconstruct masked features in the embedding space rather than the reconstruction in the input feature space. 
The predicted embeddings of masked nodes are constrained to match their representations that are directly generated from the input graph. 
Both designs naturally work as the regularization on target construction in generative graph SSL.

Inherited from GraphMAE, \model is a simple yet more effective generative self-supervised framework for graphs that can be directly coupled with existing GNN architectures. 
We perform extensive experiments on public graph datasets representative of different scales and types, including three open graph benchmark datasets. 
The results demonstrate that \model can consistently offer significant outperformance over state-of-the-art graph SSL baselines under different settings. 
Furthermore, we show that both decoding strategies contribute to the performance improvements compared to GraphMAE. 
Excitingly, \model as an SSL method offers performance advantages over classic supervised GNNs across all datasets, giving rise to the premise of self-supervised graph representation learning and pre-training.

In addition, we extend \model to large-scale graphs with hundreds of millions of nodes, which have been previously less explored for graph SSL. 
We leverage local clustering strategies that can produce local and dense subgraphs to  benefit \model (and GraphMAE) with masked feature prediction. 
Experiments on ogbn-Papers100M of 111M nodes and 1.6B edges suggest the simple \model framework can generate significant performance improvements over existing methods (Cf. Figure ~\ref{fig:head_fg}).

\section{Method}

\begin{table}[t]
    \centering
    \caption{Results with the original node features (raw) or PCA-processed node features (w/ PCA).
    \textmd{\textit{w/ PCA} represents that the input features are reduced to 50-dimensional continuous vectors using PCA, relatively less discriminative. 
     GraphMAE can be more sensitive to the discriminability of input features than the supervised one. 
    GAT is used as the backbone for all cases.}}
    \begin{tabular}{ccc}
        \toprule
                    & Cora & PubMed \\
                  &  raw $\rightarrow$ \textit{w/ PCA} &  raw $\rightarrow$ \textit{w/ PCA} \\
        \midrule
        Supervised &  83.0 $\rightarrow$ 82.3 ($\downarrow 0.7$) & 78.0 $\rightarrow$ 77.0 ($\downarrow 1.0$)  \\
         GraphMAE  &  84.2 $\rightarrow$ 82.6 ($\downarrow 1.6$) & 81.1 $\rightarrow$ 78.9 ($\downarrow 2.2$) \\
\midrule
         \model    & 84.5 $\rightarrow$ 83.5 ($\downarrow 1.0$) & 81.4 $\rightarrow$ 80.1 ($\downarrow 1.3$) \\
         \bottomrule
    \end{tabular}
    \label{tab:pre_exp}
\end{table}

\begin{figure*}[htbp]
    \centering
    \includegraphics[width=0.98\textwidth]{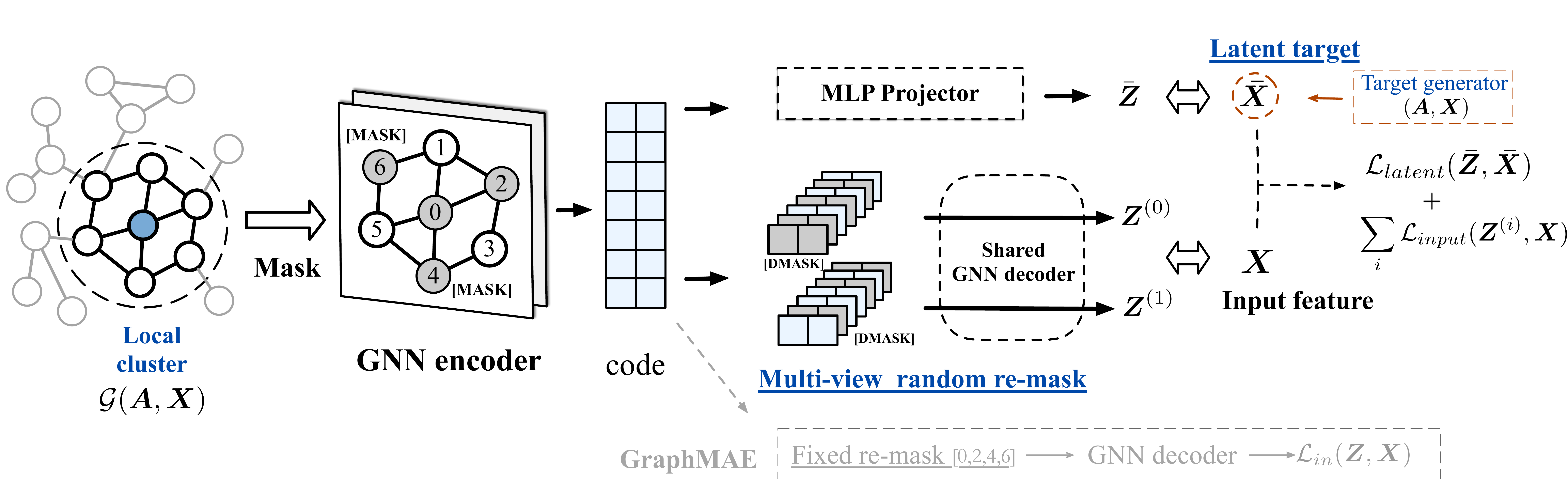}
    \caption{Overview of \model framework. \textmd{For large-scale graphs, we first run local clustering to produce local clusters for each node as the preprocessing step. During the pre-training, \model corrupts the graph by masking input node features with a mask token [MASK] and then feeds the result to a GNN encoder to generate the code. The decoding involves two objectives: 
    1) we generate multiple corrupted codes by randomly re-masking the code several times, and they are all forced to reconstruct input features after GNN decoding.
    2) we use an MLP as the decoder to predict latent target representations, which are produced by a target generator with the unmasked graph. As a comparison, GraphMAE is trained through input feature reconstruction only with a fixed re-mask decoding strategy.}}
    \label{fig:overview}
\end{figure*}

In this section, 
we first revisit masked autoencoding for graph SSL and identify its deficiency in which the effectiveness of masked feature reconstruction can be vulnerable to the distinguishability of input node features. Then we present our \model to overcome the problem by imposing regularization on the feature decoding.

\subsection{Masked Autoencoding on Graphs}

\vpara{Notations.} Let $\mathcal{G}=(\mathcal{V}, \mA, \mX)$, where $\mathcal{V}$ is the node set, $N= | \mathcal{V} | $ represents the number of nodes, $\mA \in \{0,1\}^{N \times N}$ is the adjacency matrix with each element $\mA(i,j) = 1$ indicating that there exists an edge between $v_i$ and $v_j$. $\mX \in \mathbb{R}^{N \times d_{in}}$ is the input node feature matrix. 
In graph autoencoders, we use $f_{E}$ to represent the GNN encoder such as GAT~\cite{petar2018gat} and GCN~\cite{thomas2017gcn}. And $f_{D}$ represents the decoder which can be a multi-layer perceptron (MLP) or GNN. Denoting the hidden embedding $\mH \in \mathbb{R}^{N \times d}$, the general goal of graph autoencoders is to learn representation $\mH$ or a well-initialized $f_E$ through reconstructing input node features or structure:
\begin{equation}
    \mH = f_{E}(\mA, \mX), \ \widetilde{\mathcal{G}} = f_{D}(\mA, \mH)
\end{equation}
where $\widetilde{\mathcal{G}}$ denotes the reconstructed graph characteristics, which can be structure, node features or both.

\vpara{Overview of masked feature reconstruction.} 
The idea of masked autoencoder has seen successful practice in graph SSL~\cite{hou2022graphmae}. As a form of more general denoising autoencoders, it removes a portion of data in the graph, e.g., node features or links, with the masking operation and learns to predict the masked content. 
And it has been demonstrated that reconstructing masked node features as the only pretext task could generate promising performance. 
In this work, we follow the paradigm of masked feature reconstruction and aim to further boost the performance by resolving the potential concerns in existing works. 

Formally, 
we uniformly sample a subset of nodes $\widetilde{\mathcal{V}} \subset \mathcal{V}$ without replacement and replace their feature with a mask token [MASK], i.e. a learnable vector $\vx_{[M]} \in \mathbb{R}^{d_{in}}$. 
And sampling with a relatively large mask ratio (e.g., 50\%) helps eliminate redundancy in graphs and benefit performance. The features $\widetilde{\vx}_i$ for node $v_i \in \mathcal{V}$ in the corrupted feature matrix $\widetilde{\mX}$ can be represented as:
$$
    \widetilde{\vx}_{i}=
    \begin{cases}
    \vx_{[M]} & v_i\in\widetilde{\mathcal{V}} \\
    \vx_i & v_i\notin\widetilde{\mathcal{V}} 
    \end{cases}
$$
Then the corrupted graph $(\mA, \widetilde{\mX}$) is fed into the encoder $f_E$ to generate representations $\mH$. And the decoder $f_D$ decodes the predicted masked features $\mZ$ from $\mH$. The training objective is to match the predicted $\mZ$ with the original features $\mX$ with a designated criterion, such as (scaled) cosine error.

\vpara{Problems in masked feature reconstruction.} 
Despite the excellent performance, there exists potential concern for masked node feature reconstruction due to the inaccurate semantics of node features. 
A recent study~\cite{chien2021node} shows that the performance of GNNs on downstream tasks can be significantly affected by the distinguishability of node features. In masked feature reconstruction, 
less discriminative reconstruction targets might cause misleading and harm the learning. 
To verify this assumption, we conduct pilot experiments by comparing the results using original features with less discriminative features. To induce information loss on features, we compress the features by mapping the original features to low dimensional space, i.e., 50 dimensions, using PCA. 
Table ~\ref{tab:pre_exp} shows the results. We observe that the performance of GraphMAE degrades more significantly than the supervised counterpart when using the compressed features. The results indicate that the performance of learning through 
input feature reconstruction
tends to be more vulnerable to the discriminability of the features.

In CV and NLP, where the philosophy of masked prediction has groundbreaking practices, their inputs are exact descriptions of data without loss of semantic information, e.g., pixels for images and words for texts. 
However, the input $\mX$ of graphs could inevitably 
and intrinsically 
contain unexpected noises since they processed products from various raw data, e.g., texts or hand-crafted features.
The input $\mX$ is 
and generated by various feature extractors. 
For example, the node features of Cora~\cite{yang2016revisiting} are bag-of-words vectors, ogbn-Arxiv~\cite{hu2020open} averages word embeddings of word2vec, and MAG240M~\cite{hu2021ogb} are from pretrained language model. 
Their discriminability is constrained to the expressiveness of the feature generator and could inherit the substantial noise in the generator. 
In masked feature reconstruction, the objective of recovering less discriminative node features can guide the model to fit inaccurate targets and unexpected noises, bringing potential negative effects.

\subsection{The \model Framework}
We present \model to overcome the aforementioned issue.
It follows the masked prediction paradigm and further incorporates regularization to the decoding stage to improve effectiveness.

To improve feature reconstruction, we propose to randomly re-mask the encoded representations multiple times and force the decoder to reconstruct input features from the corrupted representations. 
Then to minimize the direct effects of input features, we also enforce the model to predict representations of masked nodes in the embedding space beyond the input feature space. 
Both strategies serve as regularization to avoid the model over-fitting to the input features. 
Moreover, we extend \model to large graphs and propose to sample densely-connected subgraphs to accommodate with \model's training, 
The overall framework of \model is illustrated in Figure ~\ref{fig:overview}.

\vpara{Multi-view random re-mask decoding.} From the perspective of input feature reconstruction, we introduce randomness in the decoding and require the decoder to restore the input $\mX$ from different and partially observed embeddings. 

The decoder maps the latent code $\mH$ to the input feature space to reconstruct $\mX$ for optimization. GraphMAE~\cite{hou2022graphmae} shows that using a GNN as the decoder achieves better performance than using MLP, and the GNN decoder helps the encoder learn high-level latent code when recovering the high-dimension and low-semantic features.
The main difference is that GNN involves propagation and recovers the input relying on neighborhood information. 
Based on this characteristic of the GNN decoder, instead of the fixed re-mask decoding used in GraphMAE, we propose a \textit{multi-view random re-mask decoding} strategy. 
It randomly re-masks the encoded representation before they are fed into the decoder, which resembles the random propagation in semi-supervised learning~\cite{feng2020graph}.
Formally, 
we resample a subset of nodes $\overline{\mathcal{V}} \subset \mathcal{V}$ following a uniform distribution. $\overline{\mathcal{V}}$ is different from the input masked nodes $\widetilde{\mathcal{V}}$ and nodes are equally selected for re-masking regardless of whether they are masked before.
Then corrupted representation matrix $\widetilde{\mH}$ is built from $\mH$ by replacing the $\vh_{i}$ of node  $v_i \in \overline{\mathcal{V}}$ with another shared mask token [DMASK], i.e., a learnable vector $\vh_{[M]}\in \mathbb{R}^{d}$:
$$
    \widetilde{\vh}_{i} = 
    \begin{cases}
    \vh_{[M]} & v_i\in\overline{\mathcal{V}} \\
    \vh_i & v_i\notin\overline{\mathcal{V}} 
    \end{cases}
$$
Then the decoder would reconstruct the input $\mX$ from the corrupted $\widetilde{\mH}$.
The procedure is repeated several times to generate $K$ different re-masked nodes sets $ \{ \overline{\mathcal{V}}^{(j)} \}_{1,...,K}$ and corresponding corrupted representations $\{ \widetilde{\mH}^{(j}) \}_{1,...,K}$. Each view contains different information after re-masking, and they are all enforced to reconstruct input node features. The randomness of decoding serves as regularization preventing the network from memorizing unexpected patterns in the input $\mX$, and thus the training would be less sensitive to the disturbance in the input feature.
Finally, we employ the scaled cosine error~\cite{hou2022graphmae} to measure the reconstruction error and sum over the errors of the $K$ views for training: 
\begin{equation}
    \mathcal{L}_{input} = \frac{1}{|\widetilde{\mathcal{V}}| }\sum_{j=1}^{K}\sum_{v_i \in \widetilde{\mathcal{V}}} (1 - \frac{\vx_i^\top \vz^{(j)}_i}{\lVert \vx_i \rVert \cdot \lVert \vz_i^{(j)} \rVert })^{\gamma}
\end{equation}
where $\vx_i$ is the $i$-th row of $\mX$, $\vz_i^{(j)}$ is the $i$-th row of predicted feature $\mZ^{(j)}=f_D(\mA, \widetilde{\mH}^{(j)})$, and $\gamma >= 1$ is the scaled coefficient. 
In this work, the decoder $f_D$ for feature reconstruction consists of a light single-layer GAT. Therefore, this strategy is very efficient and only incurs negligible computational costs.

\vpara{Latent representation prediction.}
In line with the mask-then-predict, the focus of this part is on constructing an additional informative prediction target that is minimally influenced by the direct effects of input features. To achieve this, we propose to perform the prediction in representation space beyond input feature space.

Considering that the neural networks can essentially serve as denoising encoders~\cite{ma2021unified} and encode high-level semantics~\cite{zhou2021ibot, caron2021emerging}, we propose to employ a network as the target generator to produce latent prediction targets from the unmasked graph.
Formally, we denote the GNN encoder as $f_E(\cdot;\theta) = f_E$. We also define a projector $g(\cdot;\theta)$, corresponding to the decoder $f_D$ in input feature reconstruction,
to map the code $\mH$ to representation space for prediction. $\theta$ denotes their learnable weights.
The target generator network shares the same architecture as the encoder and projector but uses a different set of weights, i.e., $f_E'(\cdot; \xi)$ and $g'(\cdot; \xi)$. 
During the pretraining, the unmasked graph is first passed through the target generator to produce target representation $\bar{\mX}$. 
Then the encoding results $\mH$ of the masked graph $\mathcal{G}(\mA, \widetilde{\mX})$ are projected to representation space, resulting in $\bar{\mZ}$ for latent prediction:
\begin{equation}
      \bar{\mZ} = g(\mH; \theta), \ \bar{\mX} = g'(f_E'(\mA, \mX; \xi); \xi)
\end{equation}
The encoder and projector network are trained to match the output of the target generator on masked nodes. 
Of particular interest, encouraging the correspondence of unmasked nodes would bring slight benefits to our framework. This may attribute to the masking operation implicitly serving as a special type of augmentation. 
We learn the parameters $\theta$ of the encoder and projector by minimizing the following scaled cosine error with gradient descent. 
\begin{equation}
    \mathcal{L}_{latent} = \frac{1}{N} \sum_{i}^{N} (1 - \frac{\bar{\vz}_i^\top \bar{\vx}_i }{\lVert \bar{\vz} \rVert \cdot \lVert \bar{\vx} \rVert })^\gamma
    \label{eq:hidden_sce}
\end{equation}

And the parameters of target generator $\xi$ are updated via an exponential moving average of $\theta$ ~\cite{lillicrap2015continuous} using weight decay $\tau$:
\begin{equation}
    \xi \leftarrow  \tau \xi + (1-\tau)\theta
\end{equation}

The target generator shares similarities with the teacher network in self-knowledge distillation~\cite{caron2021emerging,zhou2021ibot} or contrastive methods~\cite{grill2020bootstrap}.
But there exist differences in both the motivation and implementation:
\model aims to direct the prediction of masked nodes with output from the unmasked graph as the target. In contrast, knowledge-distillation and contrastive methods target maximizing the consistency of two augmented views. The characteristic is that our method does not rely on any elaborate data augmentations and thus has no worry about whether the augmentations would alter the semantics in particular graphs.

\vpara{Training and inference.} The overall training flow of \model is summarized in Figure ~\ref{fig:overview}. Given a graph, the original graph is passed through the target generator to generate the latent target $\bar{\mX}$. Then we randomly mask the features of a certain portion of nodes and feed the masked graph with partially observed features into the encoder $f_E(\mA, \tilde{\mX};\theta)$ to generate the code $\mH$. Next, the decoding consists of two streams. On the one hand, we apply the multi-view random re-masking to replace re-masked nodes in $\mH$ with [DMASK] token, and the results are fed into the decoder $f_D$ to reconstruct the input $\mX$. On the other hand, another decoder $g$ is adapted to predict the latent target $\bar{\mX}$. We combine the two losses with a mixing coefficient $\lambda$ during training: 
\begin{equation}
    \mathcal{L} = \mathcal{L}_{input} + \lambda \mathcal{L}_{latent}
\end{equation}
Note that the time and space complexity of \model is linear with the number of nodes $N$, and thus it can scale to extremely large graphs.
When applying to downstream tasks, the decoder and target generator are discarded, and only the GNN encoder is used to generating embeddings or finetuned for downstream tasks.

\vpara{Extending to large-scale graph}
Extending self-supervised learning to large-scale graphs is of great practical significance, yet few efforts have been devoted to this scenario. Existing graph SSL works focus more on small graphs, and current works~\cite{hamilton2017inductive, thakoor2021bootstrapped} concerning large graphs simply conduct experiments based on existing graph sampling developed under the supervised setting, e.g., neighborhood sampling~\cite{hamilton2017inductive} or ClusterGCN~\cite{chiang2019cluster}. 
Though it is a feasible implementation, there exist several challenges that may affect the performance under the self-supervised setting: 

\begin{enumerate}[leftmargin=*]
    \item Self-supervised learning generally benefits from relatively larger model capacity, i.e., wider and deeper networks, whereas GNNs suffer from the notorious problem of over-smoothing and over-squashing when stacking more layers~\cite{li2019deepgcns,alon2020bottleneck}. One feasible way to circumvent the problems is to decouple the receptive field and depth of GNN by extracting a local subgraph ~\cite{zeng2021decoupling}. 
    \item In the context of masked feature prediction, \model  has a preference for a well-connected local structure since each node would rely on aggregating messages of its neighboring nodes to generate embedding and reconstruct features. 
\end{enumerate}

Most popular sampling methods tend to generate highly sparse yet wide subgraphs as regularization in supervised setting~\cite{zeng2020graphsaint,zou2019layer}, or only bear shallow GNNs in inference ~\cite{hamilton2017inductive,chiang2019cluster}. 
In light of these defects, we imitate the idea from ~\cite{zeng2021decoupling} and are motivated to construct densely connected subgraphs for \model to tackle the scalability on large-scale graphs. Thus, we utilize local clustering~\cite{andersen2006local,spielman2013local} algorithms to seek local and dense subgraphs. Local clustering aims to find a small cluster near a given seed in the large graph. And it has been proven to be very useful for identifying 
structures at small-scale or meso-scale~\cite{leskovec2009community,jeub2015think}. Though many local clustering algorithms have been developed, we leverage the popular spectral-based PPR-Nibble~\cite{andersen2006local} for efficient implementation. 
PPR-Nibble adopts the personalized PageRank (PPR) vector $\vp_i$, 
which reflects the significance of all nodes $\mathcal{V}$ in the graph for the node $v_i$,
to generate a local cluster for a given node $v_i$. 
Previous works~\cite{zhu2013local,yin2017local} provide a theoretical guarantee for the quality of the generated local cluster of PPR-Nibble. The theorem in~\cite{zhu2013local,yin2017local} (described in Appendix ~\ref{app:thero}) indicates that the algorithm can generate local clusters of a relatively small conductance, which meets our expectations for densely connected local subgraphs. In our work, we select the $k$-largest elements in $\vp_i$ to form a local cluster for node $v_i$ for computational efficiency. 

The PPR-Nibble can be implemented efficiently through fast approximation, and the computational complexity is linear with the number of nodes. One by-product is that this strategy decreases the discrepancy between training and inference since they are both conducted on the extracted subgraphs. 
In \model, the self-supervised learning is conducted upon all nodes within a cluster. In downstream finetuning or inference, we generate the prediction or embedding for node $v_i$ using the local cluster induced by $v_i$.

\section{Experiments}
\label{sec:experiment}
In this section, we compare our proposed self-supervised framework with state-of-the-art methods in the setting of unsupervised representation learning and semi-supervised node classification. In this work, we focus on the node classification task, which aims to predict unlabeled nodes. Note that \model is a general SSL method and can be applied to various graph learning tasks.

\begin{table}[t]
    \centering
    \caption{\label{tab:dataset}Statistics of datasets.}
    \renewcommand\tabcolsep{5.5pt}
    \begin{tabular}{l|rrr}
        \toprule[1.1pt]
        Datasets & \#Nodes  & \#Edges & \#Features \\
        \midrule 
        Cora     & 2,485  &  5,069 & 1,433  \\
        Citeseer & 2,110  & 3,668 & 3,703   \\
        Pubmed   & 19,717  & 44,324 & 300  \\
        \midrule
        ogbn-Arxiv       & 169,343 & 1,166,243 & 128 \\
        ogbn-Products    & 2,449,029 & 61,859,140 & 100 \\
        MAG-Scholar-F & 12,403,930 & 358,010,024 &  128 \\
        ogbn-Papers100M  & 111,059,956 & 1,615,685,872 & 128 \\
        \bottomrule[1.1pt]
    \end{tabular}
\end{table}

\begin{table}
    \centering
    \caption{Linear probing results on large-scale datasets with mini-batch training. \textmd{We report accuracy(\%) for all datasets. \textit{Random-Init} represents a random-initialized model without any self-supervised pretraining.}}
    \begin{threeparttable}
    \renewcommand\tabcolsep{3.5pt}
    \begin{tabular}{c|cccc}
    \toprule[1.1pt]
                & Arxiv & Products & MAG & Papers100M  \\
    \midrule
    MLP         & 55.50{\footnotesize $\pm$0.23} & 61.06{\footnotesize $\pm$0.08} & 39.11{\footnotesize $\pm$0.21} & 47.24{\footnotesize $\pm$0.31} \\
    SGC         & 66.92{\footnotesize $\pm$0.08} & 74.87{\footnotesize $\pm$0.25} & 54.68{\footnotesize $\pm$0.23} & 63.29{\footnotesize $\pm$0.19} \\
    Random-Init & 68.14{\footnotesize $\pm$0.02} & 74.04{\footnotesize $\pm$0.06} & 56.57{\footnotesize $\pm$0.03} & 61.55{\footnotesize $\pm$0.12} \\
    \midrule
    CCA-SSG     & 68.57{\footnotesize $\pm$0.02} & 75.27{\footnotesize $\pm$0.05} & 51.55{\footnotesize $\pm$0.03} & 55.67{\footnotesize $\pm$0.15} \\
    GRACE       & 69.34{\footnotesize $\pm$0.01} & \underline{79.47{\footnotesize $\pm$0.59}} & 57.39{\footnotesize $\pm$0.02} & 61.21{\footnotesize $\pm$0.12} \\
    BGRL        & 70.51{\footnotesize $\pm$0.03} & 78.59{\footnotesize $\pm$0.02} & 57.57{\footnotesize $\pm$0.01} & 62.18{\footnotesize $\pm$0.15} \\
    GGD$^{1}$         & - & 75.70{\footnotesize $\pm$0.40} & -      & \underline{63.50{\footnotesize $\pm$0.50}} \\
    GraphMAE    & \underline{71.03{\footnotesize $\pm$0.02}} & 78.89{\footnotesize $\pm$0.01} & \underline{58.75{\footnotesize $\pm$0.03}} & 62.54{\footnotesize $\pm$0.09}  \\
    \midrule
      \model &\textbf{71.89}{\footnotesize $\pm$0.03} &\textbf{81.59}{\footnotesize $\pm$0.02}  & \textbf{59.24}{\footnotesize $\pm$0.01} & \textbf{64.89}{\footnotesize $\pm$0.04}\\
    \bottomrule[1.1pt]
    \end{tabular}
            \begin{tablenotes}
            \footnotesize
            \item[1] The source code of GGD is not released and its results on Arxiv and MAG-Scholar-F are not reported in the paper.
        \end{tablenotes}
    \end{threeparttable}
    \label{tab:linearprob}
    \vspace{-2mm}
\end{table}

\subsection{Evaluating on Large-scale Datasets}
\vpara{Datasets.}
The experiments are conducted on four public datasets of different scales, varying from hundreds of thousands of nodes to hundreds of millions. The statistics are listed in Table \ref{tab:dataset}. 
In the experiments, we follow the official splits in ~\cite{hu2020open} for ogbn-Arxiv/Products/Papers100M. As for MAG-Scholar-F, we randomly select 5\%/5\%/40\% nodes for training/validation/test, respectively. 

For ogbn-Products~\cite{hu2020open} and MAG-Scholar-F~\cite{bojchevski2020scaling}, their node features are generated by first extracting bag-of-words vectors from the product descriptions or paper abstracts and then conducting Principal Component Analysis (PCA) to reduce the dimension.
ogbn-Arxiv and ogbn-Papers100M~\cite{hu2020open} are both citation networks, and they leverage word2vec model to obtain node features by averaging the embeddings of words in the paper's title and abstract.

\begin{table*}
    \centering
    \caption{Results of fine-tuning the pretrained GNN with 1\% and 5\% labeled training data on large-scale datasets. \textmd{We report accuracy(\%) for all datasets. \textit{Random-Init} represents a random-initialized model without any self-supervised pretraining.}}
    \renewcommand\tabcolsep{6.5pt}
    \begin{tabular}{c|cc|cc|cc|cc}
    \toprule[1.1pt]
                & \multicolumn{2}{c|}{Arxiv} & \multicolumn{2}{c|}{Products} & \multicolumn{2}{c|}{MAG} & \multicolumn{2}{c}{Papers100M}  \\
    \midrule
    Label ratio & 1\% &5\% &  1\% &5\% &  1\% &5\% &  1\% &5\% \\
    \midrule
    Random-Init$_{\small SAINT}$ & 63.45{\footnotesize $\pm$0.32}& 67.67{\footnotesize $\pm$0.42} & 72.23{\footnotesize $\pm$0.44} & 75.21{\footnotesize $\pm$0.35}  & 43.55{\footnotesize $\pm$0.15} & 51.03{\footnotesize $\pm$0.13} & 56.47{\footnotesize $\pm$0.23} & 60.63{\footnotesize $\pm$0.22}\\
    CCA-SSG &64.14{\footnotesize $\pm$0.21} & 68.32{\footnotesize $\pm$0.32}  & 75.89{\footnotesize $\pm$0.43} & 78.47{\footnotesize $\pm$0.42}  & 42.62{\footnotesize $\pm$0.15} & 51.32{\footnotesize $\pm$0.11} & 55.68{\footnotesize $\pm$0.24} & 59.78{\footnotesize $\pm$0.08} \\
    GRACE & 64.53{\footnotesize $\pm$0.47}& 69.21{\footnotesize $\pm$0.45}  &\textbf{77.13}{\footnotesize $\pm$0.64}  &\underline{79.67{\footnotesize $\pm$0.54}}   & 43.59{\footnotesize $\pm$0.13} &51.35{\footnotesize $\pm$0.12} &  55.45{\footnotesize $\pm$0.23} &  59.38{\footnotesize $\pm$0.15}\\
    BGRL  & 65.05{\footnotesize $\pm$1.17} &69.01{\footnotesize $\pm$0.34}  & 76.32{\footnotesize $\pm$0.54} & 79.46{\footnotesize $\pm$0.43}  & 43.92{\footnotesize $\pm$0.11} & 51.69{\footnotesize $\pm$0.15} &  55.12{\footnotesize $\pm$0.23} &  60.40{\footnotesize $\pm$0.54} \\
    \midrule
    Random-Init$_{\small LC}$ & 64.79{\footnotesize $\pm$0.45} & 67.89{\footnotesize $\pm$0.27}  & 71.87{\footnotesize $\pm$0.34} & 74.42{\footnotesize $\pm$0.43}  & 43.66{\footnotesize $\pm$0.14}& 50.86{\footnotesize $\pm$0.12}  & 57.48{\footnotesize $\pm$0.23} & 61.41{\footnotesize $\pm$0.25}  \\
      GraphMAE   &\underline{65.78{\footnotesize $\pm$0.69}} &\underline{69.78{\footnotesize $\pm$0.26}} & 75.87{\footnotesize $\pm$0.43} & 79.21{\footnotesize $\pm$0.33} &\underline{48.33{\footnotesize $\pm$0.18}}  &\underline{53.12{\footnotesize $\pm$0.12}} &\underline{58.29{\footnotesize $\pm$0.15}}  &\underline{62.00{\footnotesize $\pm$0.12}}  \\
      \model    &\textbf{66.86}{\footnotesize $\pm$0.53}  &\textbf{70.16}{\footnotesize $\pm$0.28} &\underline{76.98{\footnotesize $\pm$0.36}}  &\textbf{80.52}{\footnotesize $\pm$0.23}  &\textbf{49.01}{\footnotesize $\pm$0.15}  &\textbf{53.58}{\footnotesize $\pm$0.11} &\textbf{58.69}{\footnotesize $\pm$0.38} &\textbf{62.87}{\footnotesize $\pm$0.64}  \\
    \bottomrule[1.1pt]
    \end{tabular}
    \label{tab:finetune}
\end{table*}

\begin{table}
    \centering
    \caption{Experimental results on small-scale datasets. \textmd{We report accuracy(\%) for all datasets}. }
    \label{tab:small_data}
    \renewcommand\tabcolsep{10pt}
    \begin{tabular}{c|ccc}
        \toprule[1.1pt]
                 &   Cora      & CiteSeer      & PubMed        \\ 
         \midrule
        GCN     &  81.5          & 70.3          & 79.0           \\ 
        GAT     &  83.0{\footnotesize $\pm$0.7}  & 72.5{\footnotesize $\pm$0.7}  & 79.0{\footnotesize $\pm$0.3}     \\ 
        \midrule
        GAE     &  71.5{\footnotesize $\pm$0.4}  & 65.8{\footnotesize $\pm$0.4}  & 72.1{\footnotesize $\pm$0.5}      \\ 
        DGI     &  82.3{\footnotesize $\pm$0.6}  & 71.8{\footnotesize $\pm$0.7}  & 76.8{\footnotesize $\pm$0.6}    \\ 
        MVGRL   & 83.5{\footnotesize $\pm$0.4}   & 73.3{\footnotesize $\pm$0.5}  & 80.1{\footnotesize $\pm$0.7}      \\ 
        GRACE   & 81.9{\footnotesize $\pm$0.4}   & 71.2{\footnotesize $\pm$0.5}  & 80.6{\footnotesize $\pm$0.4}   \\ 
        BGRL    & 82.7{\footnotesize $\pm$0.6}   & 71.1{\footnotesize $\pm$0.8}  & 79.6{\footnotesize $\pm$0.5}    \\ 
        InfoGCL  & 83.5{\footnotesize $\pm$0.3}   & {\bf 73.5}{\footnotesize $\pm$0.4}  & 79.1{\footnotesize $\pm$0.2}  \\ 
        CCA-SSG & 84.0{\footnotesize $\pm$0.4}   & 73.1{\footnotesize $\pm$0.3}  & 81.0{\footnotesize $\pm$0.4} \\ 
        GGD           & 83.9{\footnotesize $\pm$0.4}  & 73.0{\footnotesize $\pm$0.6} & \underline{81.3{\footnotesize $\pm$0.8}} \\
        GraphMAE & \underline{84.2{\footnotesize $\pm$0.4}}  & \underline{73.4{\footnotesize $\pm$0.4}}  & 81.1{\footnotesize $\pm$0.4}   \\ 
        \cmidrule{1-4}
        \model  & {\bf 84.5}{\footnotesize $\pm$0.6} & \underline{73.4{\footnotesize $\pm$0.3}} & {\bf 81.4}{\footnotesize $\pm$0.5} \\ 
        \bottomrule[1.1pt]
    \end{tabular}
    \label{tab:fullbatch}
\end{table}

\vpara{Baselines.}
We compare \model with state-of-the-art self-supervised graph learning methods, including contrastive methods, GRACE~\cite{zhu2020deep}, BGRL~\cite{thakoor2021bootstrapped}, CCA-SSG~\cite{zhang2021canonical}, and GGD~\cite{zheng2022rethinking} as well as a generative method GraphMAE~\cite{hou2022graphmae}. 
Other methods are not compared because they are not scalable to large graphs, e.g., MVGRL~\cite{hassani2020contrastive}, or the source code has not been released, e.g., InfoGCL~\cite{xu2021infogcl}.
As stated in ~\cite{trivedi2022augmentations}, random models can have a strong inductive bias on graphs and are non-trivial baselines. Therefore, we also report the results of the randomly-initialized GNN model and Simplified Graph Convolution (SGC)~\cite{wu2019simplifying}, which simply stacks the propagated features of different orders, to examine whether the SSL learns a more effective propagation paradigm. Comparing with them can reflect the contributions of self-supervised learning. To extend to large graphs for baselines, we adopt GraphSAINT~\cite{zeng2020graphsaint} sampling strategy, which is proved to perform better than widely-adopted Neighborhood Sampling~\cite{hamilton2017inductive} in many cases. GraphMAE and \model are trained based on the presented local clustering algorithm. For all baselines, we employ Graph Attention Network (GAT)~\cite{petar2018gat} as the backbone of the encoder $f_E$ and the decoder for input feature reconstruction $f_D$.

\vpara{Evaluation.} \label{para:eval}
We evaluate our approach with two setups: (i) linear probing and (ii) fine-tuning. 
For \textit{linear probing}, we first generate node embeddings with the pretrained encoder. Then we discard the encoder and train a linear classifier using the embeddings under the supervised setting. 
For \textit{fine-tuning}, we add a linear classifier on top of node representations and fine-tune all parameters under the semi-supervised setting. 
We randomly sample 1\% and 5\% labels from the training set to finetune the pretrained model, aiming to test the ability to transfer knowledge learned from unlabeled data to facilitate the downstream performance with a few labels. 
For both cases, we run the experiments for 10 trials with random seeds and report the average accuracy and standard variance.

\vpara{Results.} The results of linear probing are illustrated in Table ~\ref{tab:linearprob}. And we interpret the result from 3 aspects. First, \model achieves better results than all self-supervised baselines across all datasets. This manifests that the proposed method can learn more discriminative representations under the unsupervised setting. Notably, \model improves upon GraphMAE by a margin of 1.91\% and 2.35\% (absolute difference) on MAG-Scholar-F and Papers100M. These results demonstrate the significance of the proposed improvement.
Second, our approach, together with most baselines, outperforms the randomly initialized, untrained model by a large margin. This demonstrates that the designed self-supervised pretext task guides the model to better capture the semantic and structural information than the untrained model. As a comparison, improper self-supervised signals can lead the model to perform even worse, yet this phenomenon is ignored in most previous studies. 
Third, \model consistently generates better performance than SGC. Despite the fact that methods based on decoupled propagation have achieved promising results in the full-supervised setting with the assistance of self-training, 
we demonstrate that graph neural networks, like GAT, could still be more powerful at generating node representations in the unsupervised setting. 

Table ~\ref{tab:finetune} shows the results of finetuning the pretrained model in the semi-supervised setting. On the one hand, it is observed that self-supervised pretraining of \model benefits downstream supervised training with significant performance gains. In ogbn-Products, with the pre-trained model, the performance achieves improvement by above 5.1\%.
And finetuning with only 5\% of data generates comparative performance (80.52\%) to many supervised learning methods in OGB leaderboard\footnote{\url{https://ogb.stanford.edu/docs/leader_nodeprop/}} with all 100\% of training data, e.g., GraphSAINT: 80.27, Cluster-GAT: 79.23\%. 
On the other hand, our approach remarkably achieves state-of-the-art performance for all benchmarks. The only exception is on the Products dataset with 1\% training data, where \model slightly underperforms GRACE yet still achieves the second-best result. 
It should be noted that in ogbn-Papers100M, only \model and GraphMAE generate better performance than the random-initialized model, while all contrastive baselines fail to bring improvement with pretraining. One possible reason is that the data augmentation techniques used in baselines fail in this dataset.

\subsection{Evaluating on Small-scale Datasets}
\vpara{Experimental setup.}
We also report results on small yet widely adopted datasets, i.e., Cora, Citeseer, and PubMed~\cite{yang2016revisiting}, to show the generality of our method. We follow the public data splits as ~\cite{hassani2020contrastive,velivckovic2018deep}.

We compare \model with state-of-the-art self-supervised graph learning methods, including contrastive methods, DGI~\cite{velivckovic2018deep}, MVGRL~\cite{hassani2020contrastive}, GRACE~\cite{zhu2020deep}, BGRL~\cite{thakoor2021bootstrapped},  InfoGCL~\cite{xu2021infogcl}, CCA-SSG~\cite{zhang2021canonical}, and GGD~\cite{zheng2022rethinking} as well as generative methods GAE~\cite{kipf2016variational}, GraphMAE~\cite{hou2022graphmae}. 
For the evaluation, we employ the linear probing mentioned above and report the average performance of accuracy on the test nodes based on 20 random initialization.
The GNN encoder and decoder both use standard GAT as the backbone, and an MLP is employed as the representation projector $g$.

\vpara{Results.}
From Table ~\ref{tab:small_data}, we can observe that our approach generally outperforms all baselines in all datasets, suggesting that \model serves as a general and effective framework for graph self-supervised learning on graphs of varied scales. We observe that the improvement over GraphMAE is not as significant as that in the experiments of large graphs. We guess that the reason lies in the construction of input node features. Bag-of-word vectors behave more like discrete features as words in text and pixels image and, thus are less noisy as reconstruction targets. And this may partially support our assumption that \model is more advantageous than GraphMAE when there is more noise in the data.

\begin{table}
    \centering
    \caption{Ablation studies of \model key components.}
    \renewcommand\tabcolsep{4.8pt}
    \begin{tabular}{c|ccc}
    \toprule[1.1pt]
                & Products & MAG & Papers100M \\
    \midrule
      \model & 81.59{\footnotesize$\pm$0.02}& 59.24{\footnotesize$\pm$0.01} & 64.89{\footnotesize$\pm$0.04}\\
    {\small w/o random remask} & 81.04{\footnotesize$\pm$0.03} &59.01{\footnotesize$\pm$0.02} & 64.16{\footnotesize$\pm$0.02}\\
    {\small w/o latent rep pred.} &80.01{\footnotesize$\pm$0.02} & 58.87{\footnotesize$\pm$0.02}& 62.98{\footnotesize$\pm$0.01}\\
    {\small w/o input recon.} & 76.88{\footnotesize$\pm$0.02} & 55.20{\footnotesize$\pm$0.02}&59.20{\footnotesize$\pm$0.00}\\
    {\small GraphMAE} &78.89{\footnotesize$\pm$0.01} & 58.75{\footnotesize$\pm$0.03}& 62.54{\footnotesize$\pm$0.09}\\
    \bottomrule[1.1pt]
    \end{tabular}
    \label{tab:keycomp}
\end{table}

\begin{table}
    \centering
    \caption{Ablation study on sampling strategy. \textmd{``SAINT'' refers to GraphSAINT, ``Cluster'' refers to Cluster-GCN, and ``LC'' refers the presented local clustering algorithm.}}
    \renewcommand\tabcolsep{3pt}
    \begin{tabular}{ccccc}
    \toprule[1.1pt]
               &  \small{Strategy}   & Products & MAG & Papers100M \\
    \midrule
    GRACE & {\small \textit{SAINT}} & 79.47{\footnotesize$\pm$0.59} & 57.39{\footnotesize$\pm$0.02} &61.21{\footnotesize$\pm$0.12}\\
    BGRL & {\small \textit{SAINT}} & 78.59{\footnotesize$\pm$0.02} & 57.57{\footnotesize$\pm$0.01} &62.18{\footnotesize$\pm$0.15}\\ 
    \midrule
    \model & {\small \textit{SAINT}}  &80.96{\footnotesize$\pm$0.03} & 58.75{\footnotesize$\pm$0.03}&64.21{\footnotesize$\pm$0.11}\\
    \model & {\small \textit{Cluster}} &79.35{\footnotesize$\pm$0.05} & 58.05{\footnotesize$\pm$0.02}&63.77{\footnotesize$\pm$0.11}\\
    \model & {\small \textit{LC}} & 81.59{\footnotesize$\pm$0.02} & 59.24{\footnotesize$\pm$0.01}& 64.89{\footnotesize$\pm$0.12}\\
    \bottomrule[1.1pt]
    \end{tabular}
    \label{tab:sampling}
\end{table}

\subsection{Ablation Studies}
We further conduct ablation studies to verify the contributions of the designs in \model.
We choose linear probing for evaluation.

\vpara{Ablation on the learning framework.} We study the influence of the proposed two strategies---latent representation prediction and multi-view random re-masking. The results are shown in Table ~\ref{tab:keycomp}
, where the ``w/o random re-mask'' represents that we adopt the fixed re-masking strategy as GraphMAE. 
It is observed that the two strategies both contribute to performance improvement.
This demonstrates the effectiveness and further supports our motivation for the effects of input feature quality. 
Latent representation prediction brings more benefits as the accuracy drops more when the component is removed, e.g., -1.58\% in ogbn-Products and -1.91\% in ogbn-Papers100M, than the multi-view random re-masking, e.g., -0.45\% and -0.73\%. 
The target generator network provides valuable guidance and constraints on the encoded representation.

We also conduct an experiment by totally removing the input feature reconstruction, and the training only involves latent representation prediction. 
In such cases, the learning degrades to self-knowledge distillation without heavy data augmentation and causes a significant drop in performance.
The network may fall into a trivial solution and may learn collapsed representation as the results are worse than GraphMAE or even worse than the random-initialized model in MAG-Scholar-F and ogbn-Papers100M. 
This indicates that feature reconstruction substantially supports SSL, and the proposed two strategies serve as auxiliaries to help overcome the deficiency.
Overall speaking, the results confirm that the superior performance of \model comes from the design rather than any individual contribution.

\vpara{Ablation on sampling strategies.}
Table ~\ref{tab:sampling} shows the influence of different sampling strategies.
We compare local clustering against two popular subgraph sampling algorithms--ClusterGCN~\cite{chiang2019cluster} and GraphSAINT~\cite{zeng2020graphsaint}. Neighborhood sampling is not included since
it is not friendly to masked feature reconstruction, especially with GNN decoder. The local clustering is conducive to the excellent performance of \model, as our algorithm shows an advantage over GraphSAINT and Cluster-GCN with 0.57\% and 1.49\% improvement on average. Recall that GraphSAINT tends to sample nodes globally, and thus the subgraph is more sparse. 
 Although ClusterGCN generates large and connected partitions, it suffers from high information loss as edges between clusters are abandoned. 
And the results indicate that the densely-connected local subgraph produced can generate better representations. 
In addition, we compare our approach with the strongest baselines using the same sampling strategy. And \model still generates a 1.49\% advantage in ogbn-Products, demonstrating its effectiveness.

\begin{figure}
	\centering
	\includegraphics[width=0.45\textwidth]{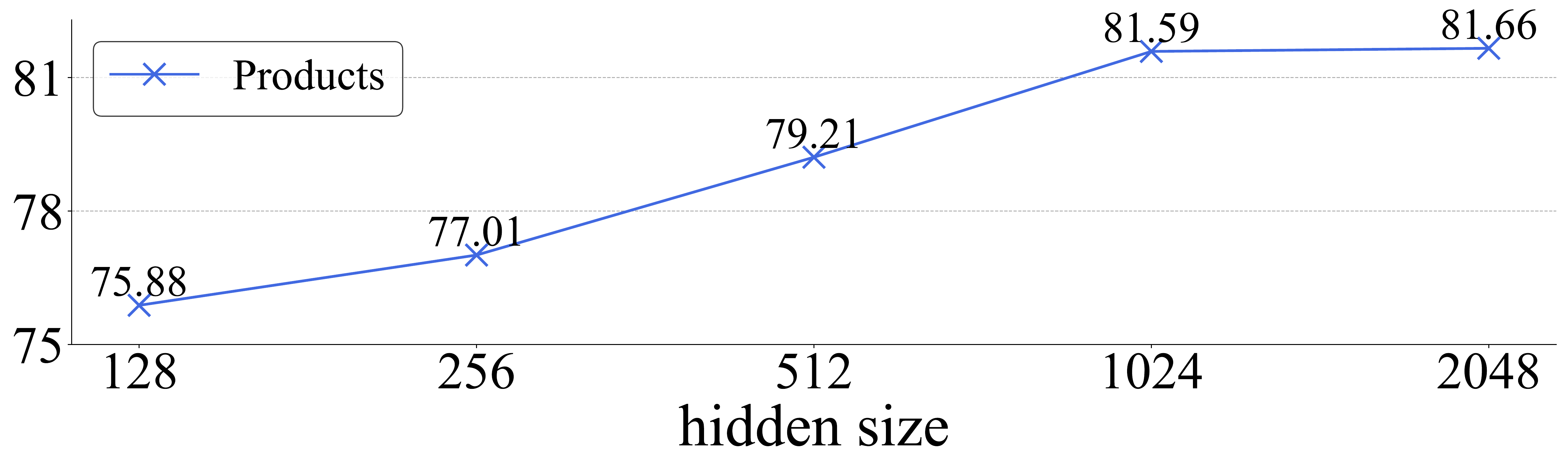} \\
	\includegraphics[width=0.47\textwidth]{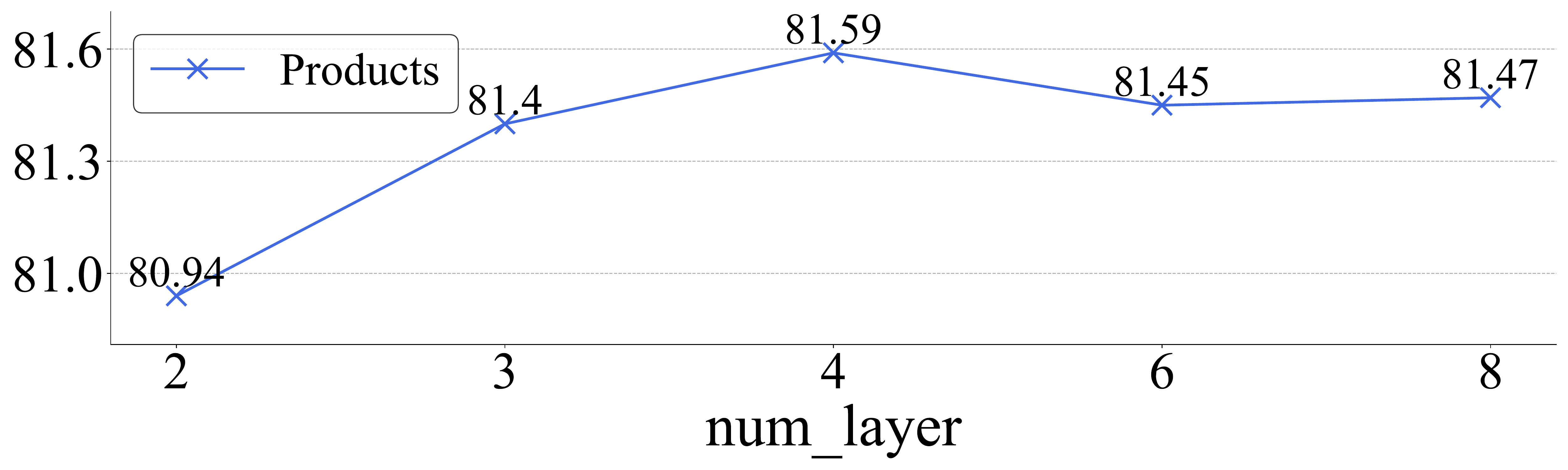}
	\caption{Ablation study on hidden size and the number of GNN layers. \textmd{The effects of width are more significant than depth.}}
	\label{fig:capacity}
\end{figure}

\vpara{Ablation on model capacity.}
The effects of model capacity have attracted significant attention in other fields like CV~\cite{he2022masked} and NLP~\cite{brown2020language} as it is demonstrated that SSL can largely benefit from increasing model parameters. 
We take an interest in whether the scaling law of model capacity also applies to GNNs. 
Specifically, we employ a GAT as the encoder and explore the influence of depth and width. And experiments are conducted on ogbn-Products of around 2 million nodes. The results are shown in Figure ~\ref{fig:capacity}. 
Increasing the hidden size drives the model to achieve better performance. Doubling the hidden size leads to a performance improvement of nearly 2\% in accuracy when the hidden size does not exceed 1024. 
But further enlarging the width only brings very marginal gain. 

Another way to increase the capacity is to stack more network layers.  
Figure~\ref{fig:capacity} shows that increasing the depth can slightly boost the performance as the accuracy increases by 0.65\% when the number of network layers is increased from 2 to 4. And the benefits would diminish when stacking more layers. One possible reason is that deeper GNNs are harder to optimize, while current downstream tasks or semantics of homogeneous structured data benefit little from more complex network architecture.
It is observed that the influence of the depth is less remarkable than the width of GNN.


\section{Related Work}

In this section, we introduce related works about graph self-supervised learning and scalable graph neural networks. 

\subsection{Graph Self-Supervised Learning}
Graph self-supervised learning (SSL) can be roughly categorized into two genres, including graph contrastive learning and graph generative learning, based on the learning paradigm. 

\vpara{Contrastive methods.}
Contrastive learning is an important way to learn representations in a self-supervised manner and has achieved successful practices in graph learning~\cite{velivckovic2018deep,sun2020infograph,hassani2020contrastive,qiu2020gcc,you2020graph,zheng2022rethinking,liu2022selfkg}. 
DGI~\cite{velivckovic2018deep} and InfoGraph~\cite{sun2020infograph} adopt the local-global mutual information maximization 
to learn node-level and graph-level representations. 
MVGRL~\cite{hassani2020contrastive} leverages graph diffusion to generate an additional view of the graph and contrasts node-graph  representations of distinct views.
GCC~\cite{qiu2020gcc} utilizes subgraph-based instance discrimination and adopts InfoNCE as the pre-training objective with MoCo-style dictionary~\cite{he2020momentum}. 
GRACE~\cite{zhu2020deep}, GraphCL~\cite{you2020graph}, and GCA~\cite{zhu2021graph} learn the node or graph representation by maximizing agreement between different augmentations. 
GGD~\cite{zheng2022rethinking} analyzes the defect of existing contrastive methods (i.e., improper usage of Sigmoid function) and proposes a group discrimination paradigm.

To avoid the expensive computation of negative samples, some researchers propose graph SSL methods that do not require negative samples. 
BGRL~\cite{thakoor2021bootstrapped} uses an online encoder and a target encoder to contrast two augmented versions without negative samples. CCA-SSG~\cite{zhang2021canonical} leverages a feature-level objective for graph SSL, inspired by Canonical Correlation Analysis methods.

Most graph contrastive learning methods rely on complex graph augmentation operators to generate two different views, which are used to be contrasted or correlated. 
However, the theoretical understanding of augmentation techniques on the graph SSL has not been well-studied. The choice of graph augmentation operators mostly depends on the empirical analysis of researchers. Although some works~\cite{xia2022simgrace,wang2022augmentation} have made attempts to alleviate this reliance, it still remains further exploration.

\vpara{Generative methods.}
Graph autoencoders (GAE) and VGAE~\cite{kipf2016variational} follow the spirit of autoencoder~\cite{hinton1993auto} to learn node representations.  
Following VGAE, most GAEs focus on reconstructing the structural information (e.g., ARVGA~\cite{pan2018adversarially}) or adopt the reconstruction objective of both structural information and node attributes (e.g., MGAE~\cite{wang2017mgae}, GALA~\cite{park2019symmetric}). 
NWR-GAE~\cite{tang2022graph} designs a graph decoder to reconstruct the entire neighborhood information of graph structure. 
kgTransformer~\cite{liu2022mask} applies the masked GAE to knowledge graph reasoning.
However, these previous GAE models do not perform well on node-level and graph-level classification tasks. 
To mitigate the performance gap, GraphMAE~\cite{hou2022graphmae} leverages masked feature reconstruction as the objective with auxiliary designs and obtains comparable or better performance than contrastive methods. 
In addition to graph autoencoders, inspired by the success of autoregressive models in natural language processing, GPT-GNN~\cite{hu2020gpt} designs an attributed graph generation task, including attribute and edge generation, for pre-training GNN models. Generative methods can alleviate the deficiency of contrastive rivals since the objective of generative ones is to directly reconstruct the input graph data.

\subsection{Scalable Graph Neural Networks}
There are two genres of methods for scalable GNNs. 
One is based on sampling that trains GNN models on sampled mini-batch data. 
GraphSAGE~\cite{hamilton2017inductive} adopts the neighbor sampling method to conduct the mini-batch training. 
FastGCN~\cite{chen2018fastgcn} performs layer-wise sampling and leverages importance sampling to reduce variance. 
GraphSAINT~\cite{zeng2020graphsaint} and ClusterGCN~\cite{chiang2019cluster} both produce a subgraph from the original graph for mini-batch training by graph partition or random walks.
Another paradigm for scalable GNNs is to decouple the message propagation and feature transformation. 
SGC~\cite{wu2019simplifying} removes the nonlinear functions and is equivalent to a pre-processing K-step propagation and a logistic regression on the propagated features. 
SIGN~\cite{frasca2020sign} extends SGC to stack the propagated results of different hops and graph filters, and then only trains MLPs for applications. 
These decoupled methods achieve excellent performance in the supervised setting but have no advantage in generating high-quality embeddings.

\section{Conclusion}
In this work, we explore graph self-supervised learning with masked feature prediction. We first examine its potential concern that the discriminability of input features hinders current attempts to achieve promising performance.
Then we present a framework \model to address this issue by imposing regularization on the prediction. We focus on the decoding stage and introduce latent representation target and randomness to input reconstruction. The novel decoding strategy significantly boosts the performance in realistic large-scale benchmarks. Our work further supports that node-level signals could provide abundant supervision for masked graph self-supervised learning and deserves further exploration.

\vpara{Acknowledgements.} This research was supported by Natural Science Foundation of China (NSFC) 61825602 and 62276148, Tsinghua-Bosch Joint ML Center, and Zhipu.AI.

\clearpage


\bibliographystyle{ACM-Reference-Format}
\bibliography{reference}


\begin{thebibliography}{60}


\ifx \showCODEN    \undefined \def \showCODEN     #1{\unskip}     \fi
\ifx \showDOI      \undefined \def \showDOI       #1{#1}\fi
\ifx \showISBNx    \undefined \def \showISBNx     #1{\unskip}     \fi
\ifx \showISBNxiii \undefined \def \showISBNxiii  #1{\unskip}     \fi
\ifx \showISSN     \undefined \def \showISSN      #1{\unskip}     \fi
\ifx \showLCCN     \undefined \def \showLCCN      #1{\unskip}     \fi
\ifx \shownote     \undefined \def \shownote      #1{#1}          \fi
\ifx \showarticletitle \undefined \def \showarticletitle #1{#1}   \fi
\ifx \showURL      \undefined \def \showURL       {\relax}        \fi
\providecommand\bibfield[2]{#2}
\providecommand\bibinfo[2]{#2}
\providecommand\natexlab[1]{#1}
\providecommand\showeprint[2][]{arXiv:#2}

\bibitem[Alon and Yahav(2020)]%
        {alon2020bottleneck}
\bibfield{author}{\bibinfo{person}{Uri Alon} {and} \bibinfo{person}{Eran
  Yahav}.} \bibinfo{year}{2020}\natexlab{}.
\newblock \showarticletitle{On the bottleneck of graph neural networks and its
  practical implications}.
\newblock \bibinfo{journal}{\emph{arXiv preprint arXiv:2006.05205}}
  (\bibinfo{year}{2020}).
\newblock


\bibitem[Andersen et~al\mbox{.}(2006)]%
        {andersen2006local}
\bibfield{author}{\bibinfo{person}{Reid Andersen}, \bibinfo{person}{Fan Chung},
  {and} \bibinfo{person}{Kevin Lang}.} \bibinfo{year}{2006}\natexlab{}.
\newblock \showarticletitle{Local graph partitioning using pagerank vectors}.
  In \bibinfo{booktitle}{\emph{FOCS}}. IEEE, \bibinfo{pages}{475--486}.
\newblock


\bibitem[Bojchevski et~al\mbox{.}(2020)]%
        {bojchevski2020scaling}
\bibfield{author}{\bibinfo{person}{Aleksandar Bojchevski},
  \bibinfo{person}{Johannes Klicpera}, \bibinfo{person}{Bryan Perozzi},
  \bibinfo{person}{Amol Kapoor}, \bibinfo{person}{Martin Blais},
  \bibinfo{person}{Benedek R{\'o}zemberczki}, \bibinfo{person}{Michal Lukasik},
  {and} \bibinfo{person}{Stephan G{\"u}nnemann}.}
  \bibinfo{year}{2020}\natexlab{}.
\newblock \showarticletitle{Scaling graph neural networks with approximate
  pagerank}. In \bibinfo{booktitle}{\emph{KDD}}. \bibinfo{pages}{2464--2473}.
\newblock


\bibitem[Brown et~al\mbox{.}(2020)]%
        {brown2020language}
\bibfield{author}{\bibinfo{person}{Tom Brown}, \bibinfo{person}{Benjamin Mann},
  \bibinfo{person}{Nick Ryder}, \bibinfo{person}{Melanie Subbiah},
  \bibinfo{person}{Jared~D Kaplan}, \bibinfo{person}{Prafulla Dhariwal},
  \bibinfo{person}{Arvind Neelakantan}, \bibinfo{person}{Pranav Shyam},
  \bibinfo{person}{Girish Sastry}, \bibinfo{person}{Amanda Askell},
  {et~al\mbox{.}}} \bibinfo{year}{2020}\natexlab{}.
\newblock \showarticletitle{Language models are few-shot learners}. In
  \bibinfo{booktitle}{\emph{NeurIPS}}, Vol.~\bibinfo{volume}{33}.
\newblock


\bibitem[Caron et~al\mbox{.}(2021)]%
        {caron2021emerging}
\bibfield{author}{\bibinfo{person}{Mathilde Caron}, \bibinfo{person}{Hugo
  Touvron}, \bibinfo{person}{Ishan Misra}, \bibinfo{person}{Herv{\'e}
  J{\'e}gou}, \bibinfo{person}{Julien Mairal}, \bibinfo{person}{Piotr
  Bojanowski}, {and} \bibinfo{person}{Armand Joulin}.}
  \bibinfo{year}{2021}\natexlab{}.
\newblock \showarticletitle{Emerging properties in self-supervised vision
  transformers}. In \bibinfo{booktitle}{\emph{ICCV}}.
  \bibinfo{pages}{9650--9660}.
\newblock


\bibitem[Chen et~al\mbox{.}(2018)]%
        {chen2018fastgcn}
\bibfield{author}{\bibinfo{person}{Jie Chen}, \bibinfo{person}{Tengfei Ma},
  {and} \bibinfo{person}{Cao Xiao}.} \bibinfo{year}{2018}\natexlab{}.
\newblock \showarticletitle{Fastgcn: fast learning with graph convolutional
  networks via importance sampling}. In \bibinfo{booktitle}{\emph{ICLR}}.
\newblock


\bibitem[Chiang et~al\mbox{.}(2019)]%
        {chiang2019cluster}
\bibfield{author}{\bibinfo{person}{Wei-Lin Chiang}, \bibinfo{person}{Xuanqing
  Liu}, \bibinfo{person}{Si Si}, \bibinfo{person}{Yang Li},
  \bibinfo{person}{Samy Bengio}, {and} \bibinfo{person}{Cho-Jui Hsieh}.}
  \bibinfo{year}{2019}\natexlab{}.
\newblock \showarticletitle{Cluster-gcn: An efficient algorithm for training
  deep and large graph convolutional networks}. In
  \bibinfo{booktitle}{\emph{KDD}}. \bibinfo{pages}{257--266}.
\newblock


\bibitem[Chien et~al\mbox{.}(2022)]%
        {chien2021node}
\bibfield{author}{\bibinfo{person}{Eli Chien}, \bibinfo{person}{Wei-Cheng
  Chang}, \bibinfo{person}{Cho-Jui Hsieh}, \bibinfo{person}{Hsiang-Fu Yu},
  \bibinfo{person}{Jiong Zhang}, \bibinfo{person}{Olgica Milenkovic}, {and}
  \bibinfo{person}{Inderjit~S Dhillon}.} \bibinfo{year}{2022}\natexlab{}.
\newblock \showarticletitle{Node Feature Extraction by Self-Supervised
  Multi-scale Neighborhood Prediction}. In \bibinfo{booktitle}{\emph{ICLR}}.
\newblock


\bibitem[Cui et~al\mbox{.}(2020)]%
        {cui2020adaptive}
\bibfield{author}{\bibinfo{person}{Ganqu Cui}, \bibinfo{person}{Jie Zhou},
  \bibinfo{person}{Cheng Yang}, {and} \bibinfo{person}{Zhiyuan Liu}.}
  \bibinfo{year}{2020}\natexlab{}.
\newblock \showarticletitle{Adaptive graph encoder for attributed graph
  embedding}. In \bibinfo{booktitle}{\emph{KDD}}. \bibinfo{pages}{976--985}.
\newblock


\bibitem[Feng et~al\mbox{.}(2020)]%
        {feng2020graph}
\bibfield{author}{\bibinfo{person}{Wenzheng Feng}, \bibinfo{person}{Jie Zhang},
  \bibinfo{person}{Yuxiao Dong}, \bibinfo{person}{Yu Han},
  \bibinfo{person}{Huanbo Luan}, \bibinfo{person}{Qian Xu},
  \bibinfo{person}{Qiang Yang}, \bibinfo{person}{Evgeny Kharlamov}, {and}
  \bibinfo{person}{Jie Tang}.} \bibinfo{year}{2020}\natexlab{}.
\newblock \showarticletitle{Graph random neural networks for semi-supervised
  learning on graphs}. In \bibinfo{booktitle}{\emph{NeurIPS}}.
\newblock


\bibitem[Frasca et~al\mbox{.}(2020)]%
        {frasca2020sign}
\bibfield{author}{\bibinfo{person}{Fabrizio Frasca}, \bibinfo{person}{Emanuele
  Rossi}, \bibinfo{person}{Davide Eynard}, \bibinfo{person}{Ben Chamberlain},
  \bibinfo{person}{Michael Bronstein}, {and} \bibinfo{person}{Federico Monti}.}
  \bibinfo{year}{2020}\natexlab{}.
\newblock \showarticletitle{Sign: Scalable inception graph neural networks}.
\newblock \bibinfo{journal}{\emph{arXiv preprint arXiv:2004.11198}}
  (\bibinfo{year}{2020}).
\newblock


\bibitem[Grill et~al\mbox{.}(2020)]%
        {grill2020bootstrap}
\bibfield{author}{\bibinfo{person}{Jean-Bastien Grill},
  \bibinfo{person}{Florian Strub}, \bibinfo{person}{Florent Altch{\'e}},
  \bibinfo{person}{Corentin Tallec}, \bibinfo{person}{Pierre~H Richemond},
  \bibinfo{person}{Elena Buchatskaya}, \bibinfo{person}{Carl Doersch},
  \bibinfo{person}{Bernardo~Avila Pires}, \bibinfo{person}{Zhaohan~Daniel Guo},
  \bibinfo{person}{Mohammad~Gheshlaghi Azar}, {et~al\mbox{.}}}
  \bibinfo{year}{2020}\natexlab{}.
\newblock \showarticletitle{Bootstrap your own latent: A new approach to
  self-supervised learning}. In \bibinfo{booktitle}{\emph{NeurIPS}}.
\newblock


\bibitem[Hamilton et~al\mbox{.}(2017)]%
        {hamilton2017inductive}
\bibfield{author}{\bibinfo{person}{Will Hamilton}, \bibinfo{person}{Zhitao
  Ying}, {and} \bibinfo{person}{Jure Leskovec}.}
  \bibinfo{year}{2017}\natexlab{}.
\newblock \showarticletitle{Inductive representation learning on large graphs}.
  In \bibinfo{booktitle}{\emph{NeurIPS}}.
\newblock


\bibitem[Hassani and Khasahmadi(2020)]%
        {hassani2020contrastive}
\bibfield{author}{\bibinfo{person}{Kaveh Hassani} {and}
  \bibinfo{person}{Amir~Hosein Khasahmadi}.} \bibinfo{year}{2020}\natexlab{}.
\newblock \showarticletitle{Contrastive multi-view representation learning on
  graphs}. In \bibinfo{booktitle}{\emph{ICML}}.
\newblock


\bibitem[He et~al\mbox{.}(2022)]%
        {he2022masked}
\bibfield{author}{\bibinfo{person}{Kaiming He}, \bibinfo{person}{Xinlei Chen},
  \bibinfo{person}{Saining Xie}, \bibinfo{person}{Yanghao Li},
  \bibinfo{person}{Piotr Doll{\'a}r}, {and} \bibinfo{person}{Ross Girshick}.}
  \bibinfo{year}{2022}\natexlab{}.
\newblock \showarticletitle{Masked autoencoders are scalable vision learners}.
  In \bibinfo{booktitle}{\emph{CVPR}}.
\newblock


\bibitem[He et~al\mbox{.}(2020)]%
        {he2020momentum}
\bibfield{author}{\bibinfo{person}{Kaiming He}, \bibinfo{person}{Haoqi Fan},
  \bibinfo{person}{Yuxin Wu}, \bibinfo{person}{Saining Xie}, {and}
  \bibinfo{person}{Ross Girshick}.} \bibinfo{year}{2020}\natexlab{}.
\newblock \showarticletitle{Momentum contrast for unsupervised visual
  representation learning}. In \bibinfo{booktitle}{\emph{CVPR}}.
\newblock


\bibitem[Hinton and Zemel(1993)]%
        {hinton1993auto}
\bibfield{author}{\bibinfo{person}{Geoffrey~E Hinton} {and}
  \bibinfo{person}{Richard Zemel}.} \bibinfo{year}{1993}\natexlab{}.
\newblock \showarticletitle{Autoencoders, Minimum Description Length and
  Helmholtz Free Energy}. In \bibinfo{booktitle}{\emph{NeurIPS}},
  \bibfield{editor}{\bibinfo{person}{J.~Cowan}, \bibinfo{person}{G.~Tesauro},
  {and} \bibinfo{person}{J.~Alspector}} (Eds.), Vol.~\bibinfo{volume}{6}.
  \bibinfo{publisher}{Morgan-Kaufmann}.
\newblock


\bibitem[Hou et~al\mbox{.}(2022)]%
        {hou2022graphmae}
\bibfield{author}{\bibinfo{person}{Zhenyu Hou}, \bibinfo{person}{Xiao Liu},
  \bibinfo{person}{Yukuo Cen}, \bibinfo{person}{Yuxiao Dong},
  \bibinfo{person}{Hongxia Yang}, \bibinfo{person}{Chunjie Wang}, {and}
  \bibinfo{person}{Jie Tang}.} \bibinfo{year}{2022}\natexlab{}.
\newblock \showarticletitle{GraphMAE: Self-Supervised Masked Graph
  Autoencoders}. In \bibinfo{booktitle}{\emph{KDD}}.
\newblock


\bibitem[Hu et~al\mbox{.}(2021)]%
        {hu2021ogb}
\bibfield{author}{\bibinfo{person}{Weihua Hu}, \bibinfo{person}{Matthias Fey},
  \bibinfo{person}{Hongyu Ren}, \bibinfo{person}{Maho Nakata},
  \bibinfo{person}{Yuxiao Dong}, {and} \bibinfo{person}{Jure Leskovec}.}
  \bibinfo{year}{2021}\natexlab{}.
\newblock \showarticletitle{Ogb-lsc: A large-scale challenge for machine
  learning on graphs}.
\newblock \bibinfo{journal}{\emph{arXiv preprint arXiv:2103.09430}}
  (\bibinfo{year}{2021}).
\newblock


\bibitem[Hu et~al\mbox{.}(2020b)]%
        {hu2020open}
\bibfield{author}{\bibinfo{person}{Weihua Hu}, \bibinfo{person}{Matthias Fey},
  \bibinfo{person}{Marinka Zitnik}, \bibinfo{person}{Yuxiao Dong},
  \bibinfo{person}{Hongyu Ren}, \bibinfo{person}{Bowen Liu},
  \bibinfo{person}{Michele Catasta}, {and} \bibinfo{person}{Jure Leskovec}.}
  \bibinfo{year}{2020}\natexlab{b}.
\newblock \showarticletitle{Open graph benchmark: Datasets for machine learning
  on graphs}. In \bibinfo{booktitle}{\emph{NeurIPS}}.
\newblock


\bibitem[Hu et~al\mbox{.}(2019)]%
        {hu2019strategies}
\bibfield{author}{\bibinfo{person}{Weihua Hu}, \bibinfo{person}{Bowen Liu},
  \bibinfo{person}{Joseph Gomes}, \bibinfo{person}{Marinka Zitnik},
  \bibinfo{person}{Percy Liang}, \bibinfo{person}{Vijay Pande}, {and}
  \bibinfo{person}{Jure Leskovec}.} \bibinfo{year}{2019}\natexlab{}.
\newblock \showarticletitle{Strategies for pre-training graph neural networks}.
  In \bibinfo{booktitle}{\emph{ICLR}}.
\newblock


\bibitem[Hu et~al\mbox{.}(2020a)]%
        {hu2020gpt}
\bibfield{author}{\bibinfo{person}{Ziniu Hu}, \bibinfo{person}{Yuxiao Dong},
  \bibinfo{person}{Kuansan Wang}, \bibinfo{person}{Kai-Wei Chang}, {and}
  \bibinfo{person}{Yizhou Sun}.} \bibinfo{year}{2020}\natexlab{a}.
\newblock \showarticletitle{Gpt-gnn: Generative pre-training of graph neural
  networks}. In \bibinfo{booktitle}{\emph{KDD}}.
\newblock


\bibitem[Jeub et~al\mbox{.}(2015)]%
        {jeub2015think}
\bibfield{author}{\bibinfo{person}{Lucas~GS Jeub}, \bibinfo{person}{Prakash
  Balachandran}, \bibinfo{person}{Mason~A Porter}, \bibinfo{person}{Peter~J
  Mucha}, {and} \bibinfo{person}{Michael~W Mahoney}.}
  \bibinfo{year}{2015}\natexlab{}.
\newblock \showarticletitle{Think locally, act locally: Detection of small,
  medium-sized, and large communities in large networks}.
\newblock \bibinfo{journal}{\emph{Physical Review E}} \bibinfo{volume}{91},
  \bibinfo{number}{1} (\bibinfo{year}{2015}).
\newblock


\bibitem[Kipf and Welling(2016)]%
        {kipf2016variational}
\bibfield{author}{\bibinfo{person}{Thomas~N Kipf} {and} \bibinfo{person}{Max
  Welling}.} \bibinfo{year}{2016}\natexlab{}.
\newblock \showarticletitle{Variational graph auto-encoders}.
\newblock \bibinfo{journal}{\emph{arXiv preprint arXiv:1611.07308}}
  (\bibinfo{year}{2016}).
\newblock


\bibitem[Kipf and Welling(2017)]%
        {thomas2017gcn}
\bibfield{author}{\bibinfo{person}{Thomas~N. Kipf} {and} \bibinfo{person}{Max
  Welling}.} \bibinfo{year}{2017}\natexlab{}.
\newblock \showarticletitle{Semi-Supervised Classification with Graph
  Convolutional Networks}. In \bibinfo{booktitle}{\emph{ICLR}}.
\newblock


\bibitem[Leskovec et~al\mbox{.}(2009)]%
        {leskovec2009community}
\bibfield{author}{\bibinfo{person}{Jure Leskovec}, \bibinfo{person}{Kevin~J
  Lang}, \bibinfo{person}{Anirban Dasgupta}, {and} \bibinfo{person}{Michael~W
  Mahoney}.} \bibinfo{year}{2009}\natexlab{}.
\newblock \showarticletitle{Community structure in large networks: Natural
  cluster sizes and the absence of large well-defined clusters}.
\newblock \bibinfo{journal}{\emph{Internet Mathematics}} \bibinfo{volume}{6},
  \bibinfo{number}{1} (\bibinfo{year}{2009}), \bibinfo{pages}{29--123}.
\newblock


\bibitem[Li et~al\mbox{.}(2019)]%
        {li2019deepgcns}
\bibfield{author}{\bibinfo{person}{Guohao Li}, \bibinfo{person}{Matthias
  Muller}, \bibinfo{person}{Ali Thabet}, {and} \bibinfo{person}{Bernard
  Ghanem}.} \bibinfo{year}{2019}\natexlab{}.
\newblock \showarticletitle{Deepgcns: Can gcns go as deep as cnns?}. In
  \bibinfo{booktitle}{\emph{ICCV}}. \bibinfo{pages}{9267--9276}.
\newblock


\bibitem[Lillicrap et~al\mbox{.}(2015)]%
        {lillicrap2015continuous}
\bibfield{author}{\bibinfo{person}{Timothy~P Lillicrap},
  \bibinfo{person}{Jonathan~J Hunt}, \bibinfo{person}{Alexander Pritzel},
  \bibinfo{person}{Nicolas Heess}, \bibinfo{person}{Tom Erez},
  \bibinfo{person}{Yuval Tassa}, \bibinfo{person}{David Silver}, {and}
  \bibinfo{person}{Daan Wierstra}.} \bibinfo{year}{2015}\natexlab{}.
\newblock \showarticletitle{Continuous control with deep reinforcement
  learning}.
\newblock \bibinfo{journal}{\emph{arXiv preprint arXiv:1509.02971}}
  (\bibinfo{year}{2015}).
\newblock


\bibitem[Liu et~al\mbox{.}(2022a)]%
        {liu2022selfkg}
\bibfield{author}{\bibinfo{person}{Xiao Liu}, \bibinfo{person}{Haoyun Hong},
  \bibinfo{person}{Xinghao Wang}, \bibinfo{person}{Zeyi Chen},
  \bibinfo{person}{Evgeny Kharlamov}, \bibinfo{person}{Yuxiao Dong}, {and}
  \bibinfo{person}{Jie Tang}.} \bibinfo{year}{2022}\natexlab{a}.
\newblock \showarticletitle{Selfkg: self-supervised entity alignment in
  knowledge graphs}. In \bibinfo{booktitle}{\emph{WWW}}.
  \bibinfo{pages}{860--870}.
\newblock


\bibitem[Liu et~al\mbox{.}(2021)]%
        {liu2021self}
\bibfield{author}{\bibinfo{person}{Xiao Liu}, \bibinfo{person}{Fanjin Zhang},
  \bibinfo{person}{Zhenyu Hou}, \bibinfo{person}{Li Mian},
  \bibinfo{person}{Zhaoyu Wang}, \bibinfo{person}{Jing Zhang}, {and}
  \bibinfo{person}{Jie Tang}.} \bibinfo{year}{2021}\natexlab{}.
\newblock \showarticletitle{Self-supervised learning: Generative or
  contrastive}.
\newblock \bibinfo{journal}{\emph{TKDE}} (\bibinfo{year}{2021}).
\newblock


\bibitem[Liu et~al\mbox{.}(2022b)]%
        {liu2022mask}
\bibfield{author}{\bibinfo{person}{Xiao Liu}, \bibinfo{person}{Shiyu Zhao},
  \bibinfo{person}{Kai Su}, \bibinfo{person}{Yukuo Cen},
  \bibinfo{person}{Jiezhong Qiu}, \bibinfo{person}{Mengdi Zhang},
  \bibinfo{person}{Wei Wu}, \bibinfo{person}{Yuxiao Dong}, {and}
  \bibinfo{person}{Jie Tang}.} \bibinfo{year}{2022}\natexlab{b}.
\newblock \showarticletitle{Mask and Reason: Pre-Training Knowledge Graph
  Transformers for Complex Logical Queries}. In
  \bibinfo{booktitle}{\emph{KDD}}. \bibinfo{pages}{1120--1130}.
\newblock


\bibitem[Ma et~al\mbox{.}(2021)]%
        {ma2021unified}
\bibfield{author}{\bibinfo{person}{Yao Ma}, \bibinfo{person}{Xiaorui Liu},
  \bibinfo{person}{Tong Zhao}, \bibinfo{person}{Yozen Liu},
  \bibinfo{person}{Jiliang Tang}, {and} \bibinfo{person}{Neil Shah}.}
  \bibinfo{year}{2021}\natexlab{}.
\newblock \showarticletitle{A unified view on graph neural networks as graph
  signal denoising}. In \bibinfo{booktitle}{\emph{CIKM}}.
\newblock


\bibitem[Pan et~al\mbox{.}(2018)]%
        {pan2018adversarially}
\bibfield{author}{\bibinfo{person}{Shirui Pan}, \bibinfo{person}{Ruiqi Hu},
  \bibinfo{person}{Guodong Long}, \bibinfo{person}{Jing Jiang},
  \bibinfo{person}{Lina Yao}, {and} \bibinfo{person}{Chengqi Zhang}.}
  \bibinfo{year}{2018}\natexlab{}.
\newblock \showarticletitle{Adversarially regularized graph autoencoder for
  graph embedding}. In \bibinfo{booktitle}{\emph{IJCAI}}.
\newblock


\bibitem[Park et~al\mbox{.}(2019)]%
        {park2019symmetric}
\bibfield{author}{\bibinfo{person}{Jiwoong Park}, \bibinfo{person}{Minsik Lee},
  \bibinfo{person}{Hyung~Jin Chang}, \bibinfo{person}{Kyuewang Lee}, {and}
  \bibinfo{person}{Jin~Young Choi}.} \bibinfo{year}{2019}\natexlab{}.
\newblock \showarticletitle{Symmetric graph convolutional autoencoder for
  unsupervised graph representation learning}. In
  \bibinfo{booktitle}{\emph{ICCV}}. \bibinfo{pages}{6519--6528}.
\newblock


\bibitem[Qiu et~al\mbox{.}(2020)]%
        {qiu2020gcc}
\bibfield{author}{\bibinfo{person}{Jiezhong Qiu}, \bibinfo{person}{Qibin Chen},
  \bibinfo{person}{Yuxiao Dong}, \bibinfo{person}{Jing Zhang},
  \bibinfo{person}{Hongxia Yang}, \bibinfo{person}{Ming Ding},
  \bibinfo{person}{Kuansan Wang}, {and} \bibinfo{person}{Jie Tang}.}
  \bibinfo{year}{2020}\natexlab{}.
\newblock \showarticletitle{Gcc: Graph contrastive coding for graph neural
  network pre-training}. In \bibinfo{booktitle}{\emph{SIGKDD}}.
\newblock


\bibitem[Spielman and Teng(2013)]%
        {spielman2013local}
\bibfield{author}{\bibinfo{person}{Daniel~A Spielman} {and}
  \bibinfo{person}{Shang-Hua Teng}.} \bibinfo{year}{2013}\natexlab{}.
\newblock \showarticletitle{A local clustering algorithm for massive graphs and
  its application to nearly linear time graph partitioning}.
\newblock \bibinfo{journal}{\emph{SIAM Journal on computing}}
  \bibinfo{volume}{42}, \bibinfo{number}{1} (\bibinfo{year}{2013}),
  \bibinfo{pages}{1--26}.
\newblock


\bibitem[Sun et~al\mbox{.}(2020)]%
        {sun2020infograph}
\bibfield{author}{\bibinfo{person}{Fan-Yun Sun}, \bibinfo{person}{Jordan
  Hoffmann}, \bibinfo{person}{Vikas Verma}, {and} \bibinfo{person}{Jian Tang}.}
  \bibinfo{year}{2020}\natexlab{}.
\newblock \showarticletitle{Infograph: Unsupervised and semi-supervised
  graph-level representation learning via mutual information maximization}. In
  \bibinfo{booktitle}{\emph{ICLR'20}}.
\newblock


\bibitem[Tang et~al\mbox{.}(2022)]%
        {tang2022graph}
\bibfield{author}{\bibinfo{person}{Mingyue Tang}, \bibinfo{person}{Carl Yang},
  {and} \bibinfo{person}{Pan Li}.} \bibinfo{year}{2022}\natexlab{}.
\newblock \showarticletitle{Graph Auto-Encoder via Neighborhood Wasserstein
  Reconstruction}.
\newblock \bibinfo{journal}{\emph{arXiv preprint arXiv:2202.09025}}
  (\bibinfo{year}{2022}).
\newblock


\bibitem[Thakoor et~al\mbox{.}(2022)]%
        {thakoor2021bootstrapped}
\bibfield{author}{\bibinfo{person}{Shantanu Thakoor}, \bibinfo{person}{Corentin
  Tallec}, \bibinfo{person}{Mohammad~Gheshlaghi Azar},
  \bibinfo{person}{R{\'e}mi Munos}, \bibinfo{person}{Petar
  Veli{\v{c}}kovi{\'c}}, {and} \bibinfo{person}{Michal Valko}.}
  \bibinfo{year}{2022}\natexlab{}.
\newblock \showarticletitle{Large-Scale Representation Learning on Graphs via
  Bootstrapping}. In \bibinfo{booktitle}{\emph{ICLR}}.
\newblock


\bibitem[Trivedi et~al\mbox{.}(2022)]%
        {trivedi2022augmentations}
\bibfield{author}{\bibinfo{person}{Puja Trivedi}, \bibinfo{person}{Ekdeep~Singh
  Lubana}, \bibinfo{person}{Yujun Yan}, \bibinfo{person}{Yaoqing Yang}, {and}
  \bibinfo{person}{Danai Koutra}.} \bibinfo{year}{2022}\natexlab{}.
\newblock \showarticletitle{Augmentations in graph contrastive learning:
  Current methodological flaws \& towards better practices}. In
  \bibinfo{booktitle}{\emph{WWW}}. \bibinfo{pages}{1538--1549}.
\newblock


\bibitem[Velickovic et~al\mbox{.}(2018)]%
        {petar2018gat}
\bibfield{author}{\bibinfo{person}{Petar Velickovic}, \bibinfo{person}{Guillem
  Cucurull}, \bibinfo{person}{Arantxa Casanova}, \bibinfo{person}{Adriana
  Romero}, \bibinfo{person}{Pietro Li{\`{o}}}, {and} \bibinfo{person}{Yoshua
  Bengio}.} \bibinfo{year}{2018}\natexlab{}.
\newblock \showarticletitle{Graph Attention Networks}. In
  \bibinfo{booktitle}{\emph{ICLR}}.
\newblock


\bibitem[Veli{\v{c}}kovi{\'c} et~al\mbox{.}(2018)]%
        {velivckovic2018deep}
\bibfield{author}{\bibinfo{person}{Petar Veli{\v{c}}kovi{\'c}},
  \bibinfo{person}{William Fedus}, \bibinfo{person}{William~L Hamilton},
  \bibinfo{person}{Pietro Li{\`o}}, \bibinfo{person}{Yoshua Bengio}, {and}
  \bibinfo{person}{R~Devon Hjelm}.} \bibinfo{year}{2018}\natexlab{}.
\newblock \showarticletitle{Deep Graph Infomax}. In
  \bibinfo{booktitle}{\emph{ICLR}}.
\newblock


\bibitem[Wang et~al\mbox{.}(2017)]%
        {wang2017mgae}
\bibfield{author}{\bibinfo{person}{Chun Wang}, \bibinfo{person}{Shirui Pan},
  \bibinfo{person}{Guodong Long}, \bibinfo{person}{Xingquan Zhu}, {and}
  \bibinfo{person}{Jing Jiang}.} \bibinfo{year}{2017}\natexlab{}.
\newblock \showarticletitle{Mgae: Marginalized graph autoencoder for graph
  clustering}. In \bibinfo{booktitle}{\emph{CIKM}}. \bibinfo{pages}{889--898}.
\newblock


\bibitem[Wang et~al\mbox{.}(2022)]%
        {wang2022augmentation}
\bibfield{author}{\bibinfo{person}{Haonan Wang}, \bibinfo{person}{Jieyu Zhang},
  \bibinfo{person}{Qi Zhu}, {and} \bibinfo{person}{Wei Huang}.}
  \bibinfo{year}{2022}\natexlab{}.
\newblock \showarticletitle{Augmentation-Free Graph Contrastive Learning}. In
  \bibinfo{booktitle}{\emph{AAAI}}.
\newblock


\bibitem[Wei et~al\mbox{.}(2022)]%
        {wei2022masked}
\bibfield{author}{\bibinfo{person}{Chen Wei}, \bibinfo{person}{Haoqi Fan},
  \bibinfo{person}{Saining Xie}, \bibinfo{person}{Chao-Yuan Wu},
  \bibinfo{person}{Alan Yuille}, {and} \bibinfo{person}{Christoph
  Feichtenhofer}.} \bibinfo{year}{2022}\natexlab{}.
\newblock \showarticletitle{Masked feature prediction for self-supervised
  visual pre-training}. In \bibinfo{booktitle}{\emph{CVPR}}.
  \bibinfo{pages}{14668--14678}.
\newblock


\bibitem[Wu et~al\mbox{.}(2019)]%
        {wu2019simplifying}
\bibfield{author}{\bibinfo{person}{Felix Wu}, \bibinfo{person}{Amauri Souza},
  \bibinfo{person}{Tianyi Zhang}, \bibinfo{person}{Christopher Fifty},
  \bibinfo{person}{Tao Yu}, {and} \bibinfo{person}{Kilian Weinberger}.}
  \bibinfo{year}{2019}\natexlab{}.
\newblock \showarticletitle{Simplifying graph convolutional networks}. In
  \bibinfo{booktitle}{\emph{ICML}}. PMLR.
\newblock


\bibitem[Xia et~al\mbox{.}(2022)]%
        {xia2022simgrace}
\bibfield{author}{\bibinfo{person}{Jun Xia}, \bibinfo{person}{Lirong Wu},
  \bibinfo{person}{Jintao Chen}, \bibinfo{person}{Bozhen Hu}, {and}
  \bibinfo{person}{Stan~Z Li}.} \bibinfo{year}{2022}\natexlab{}.
\newblock \showarticletitle{SimGRACE: A Simple Framework for Graph Contrastive
  Learning without Data Augmentation}. In \bibinfo{booktitle}{\emph{WWW}}.
  \bibinfo{pages}{1070--1079}.
\newblock


\bibitem[Xu et~al\mbox{.}(2021)]%
        {xu2021infogcl}
\bibfield{author}{\bibinfo{person}{Dongkuan Xu}, \bibinfo{person}{Wei Cheng},
  \bibinfo{person}{Dongsheng Luo}, \bibinfo{person}{Haifeng Chen}, {and}
  \bibinfo{person}{Xiang Zhang}.} \bibinfo{year}{2021}\natexlab{}.
\newblock \showarticletitle{Infogcl: Information-aware graph contrastive
  learning}.
\newblock \bibinfo{journal}{\emph{NeurIPS}}  \bibinfo{volume}{34}
  (\bibinfo{year}{2021}).
\newblock


\bibitem[Yang et~al\mbox{.}(2016)]%
        {yang2016revisiting}
\bibfield{author}{\bibinfo{person}{Zhilin Yang}, \bibinfo{person}{William
  Cohen}, {and} \bibinfo{person}{Ruslan Salakhudinov}.}
  \bibinfo{year}{2016}\natexlab{}.
\newblock \showarticletitle{Revisiting semi-supervised learning with graph
  embeddings}. In \bibinfo{booktitle}{\emph{ICML}}. PMLR,
  \bibinfo{pages}{40--48}.
\newblock


\bibitem[Yin et~al\mbox{.}(2017)]%
        {yin2017local}
\bibfield{author}{\bibinfo{person}{Hao Yin}, \bibinfo{person}{Austin~R Benson},
  \bibinfo{person}{Jure Leskovec}, {and} \bibinfo{person}{David~F Gleich}.}
  \bibinfo{year}{2017}\natexlab{}.
\newblock \showarticletitle{Local higher-order graph clustering}. In
  \bibinfo{booktitle}{\emph{KDD}}. \bibinfo{pages}{555--564}.
\newblock


\bibitem[You et~al\mbox{.}(2020)]%
        {you2020graph}
\bibfield{author}{\bibinfo{person}{Yuning You}, \bibinfo{person}{Tianlong
  Chen}, \bibinfo{person}{Yongduo Sui}, \bibinfo{person}{Ting Chen},
  \bibinfo{person}{Zhangyang Wang}, {and} \bibinfo{person}{Yang Shen}.}
  \bibinfo{year}{2020}\natexlab{}.
\newblock \showarticletitle{Graph contrastive learning with augmentations}. In
  \bibinfo{booktitle}{\emph{NeurIPS}}.
\newblock


\bibitem[Zeng et~al\mbox{.}(2021)]%
        {zeng2021decoupling}
\bibfield{author}{\bibinfo{person}{Hanqing Zeng}, \bibinfo{person}{Muhan
  Zhang}, \bibinfo{person}{Yinglong Xia}, \bibinfo{person}{Ajitesh Srivastava},
  \bibinfo{person}{Andrey Malevich}, \bibinfo{person}{Rajgopal Kannan},
  \bibinfo{person}{Viktor Prasanna}, \bibinfo{person}{Long Jin}, {and}
  \bibinfo{person}{Ren Chen}.} \bibinfo{year}{2021}\natexlab{}.
\newblock \showarticletitle{Decoupling the depth and scope of graph neural
  networks}. In \bibinfo{booktitle}{\emph{NeurIPS}}.
\newblock


\bibitem[Zeng et~al\mbox{.}(2020)]%
        {zeng2020graphsaint}
\bibfield{author}{\bibinfo{person}{Hanqing Zeng}, \bibinfo{person}{Hongkuan
  Zhou}, \bibinfo{person}{Ajitesh Srivastava}, \bibinfo{person}{Rajgopal
  Kannan}, {and} \bibinfo{person}{Viktor Prasanna}.}
  \bibinfo{year}{2020}\natexlab{}.
\newblock \showarticletitle{Graphsaint: Graph sampling based inductive learning
  method}. In \bibinfo{booktitle}{\emph{ICLR}}.
\newblock


\bibitem[Zhang et~al\mbox{.}(2021)]%
        {zhang2021canonical}
\bibfield{author}{\bibinfo{person}{Hengrui Zhang}, \bibinfo{person}{Qitian Wu},
  \bibinfo{person}{Junchi Yan}, \bibinfo{person}{David Wipf}, {and}
  \bibinfo{person}{Philip~S Yu}.} \bibinfo{year}{2021}\natexlab{}.
\newblock \showarticletitle{From canonical correlation analysis to
  self-supervised graph neural networks}. In
  \bibinfo{booktitle}{\emph{NeurIPS}}.
\newblock


\bibitem[Zheng et~al\mbox{.}(2022)]%
        {zheng2022rethinking}
\bibfield{author}{\bibinfo{person}{Yizhen Zheng}, \bibinfo{person}{Shirui Pan},
  \bibinfo{person}{Vincent~Cs Lee}, \bibinfo{person}{Yu Zheng}, {and}
  \bibinfo{person}{Philip~S Yu}.} \bibinfo{year}{2022}\natexlab{}.
\newblock \showarticletitle{Rethinking and Scaling Up Graph Contrastive
  Learning: An Extremely Efficient Approach with Group Discrimination}. In
  \bibinfo{booktitle}{\emph{NeurIPS}}.
\newblock


\bibitem[Zhou et~al\mbox{.}(2022)]%
        {zhou2021ibot}
\bibfield{author}{\bibinfo{person}{Jinghao Zhou}, \bibinfo{person}{Chen Wei},
  \bibinfo{person}{Huiyu Wang}, \bibinfo{person}{Wei Shen},
  \bibinfo{person}{Cihang Xie}, \bibinfo{person}{Alan Yuille}, {and}
  \bibinfo{person}{Tao Kong}.} \bibinfo{year}{2022}\natexlab{}.
\newblock \showarticletitle{ibot: Image bert pre-training with online
  tokenizer}. In \bibinfo{booktitle}{\emph{ICLR}}.
\newblock


\bibitem[Zhu et~al\mbox{.}(2020)]%
        {zhu2020deep}
\bibfield{author}{\bibinfo{person}{Yanqiao Zhu}, \bibinfo{person}{Yichen Xu},
  \bibinfo{person}{Feng Yu}, \bibinfo{person}{Qiang Liu}, \bibinfo{person}{Shu
  Wu}, {and} \bibinfo{person}{Liang Wang}.} \bibinfo{year}{2020}\natexlab{}.
\newblock \showarticletitle{Deep graph contrastive representation learning}.
\newblock \bibinfo{journal}{\emph{arXiv preprint arXiv:2006.04131}}
  (\bibinfo{year}{2020}).
\newblock


\bibitem[Zhu et~al\mbox{.}(2021)]%
        {zhu2021graph}
\bibfield{author}{\bibinfo{person}{Yanqiao Zhu}, \bibinfo{person}{Yichen Xu},
  \bibinfo{person}{Feng Yu}, \bibinfo{person}{Qiang Liu}, \bibinfo{person}{Shu
  Wu}, {and} \bibinfo{person}{Liang Wang}.} \bibinfo{year}{2021}\natexlab{}.
\newblock \showarticletitle{Graph contrastive learning with adaptive
  augmentation}. In \bibinfo{booktitle}{\emph{WWW}}.
  \bibinfo{pages}{2069--2080}.
\newblock


\bibitem[Zhu et~al\mbox{.}(2013)]%
        {zhu2013local}
\bibfield{author}{\bibinfo{person}{Zeyuan~Allen Zhu}, \bibinfo{person}{Silvio
  Lattanzi}, {and} \bibinfo{person}{Vahab Mirrokni}.}
  \bibinfo{year}{2013}\natexlab{}.
\newblock \showarticletitle{A local algorithm for finding well-connected
  clusters}. In \bibinfo{booktitle}{\emph{ICML}}. PMLR,
  \bibinfo{pages}{396--404}.
\newblock


\bibitem[Zou et~al\mbox{.}(2019)]%
        {zou2019layer}
\bibfield{author}{\bibinfo{person}{Difan Zou}, \bibinfo{person}{Ziniu Hu},
  \bibinfo{person}{Yewen Wang}, \bibinfo{person}{Song Jiang},
  \bibinfo{person}{Yizhou Sun}, {and} \bibinfo{person}{Quanquan Gu}.}
  \bibinfo{year}{2019}\natexlab{}.
\newblock \showarticletitle{Layer-dependent importance sampling for training
  deep and large graph convolutional networks}. In
  \bibinfo{booktitle}{\emph{NeurIPS}}.
\newblock


\end{thebibliography}

\clearpage

\appendix
\section{Appendix}
\label{sec:appendix}

\subsection{Theorem for Local Clustering.}
\label{app:thero}
\begin{theorem}[Theorem 1 in \cite{zhu2013local}; Theorem 4.3 in \cite{yin2017local}]
\label{thm:nibble}
Let $T \subset V$ be some unknown targeted cluster
that we are trying to retrieve from an unweighted graph.
Let $\eta$ be the inverse mixing time of the random walk on the subgraph
induced by $T$.
Then there exists $T^g \subseteq T$
with $\vol{(T^g)}\geq \vol{(T)}/2$, such that
for any seed $u \in T^g$, 
PPR-Nibble with $\alpha = \Theta(\eta)$ and 
$\epsilon \in \left[\frac{1}{10\vol{(T)}}, \frac{1}{5\vol{(T)}} \right]$ outputs a set $S$ with
\begin{equation}
\nonumber
\Phi(S) \leq \widetilde{O}\left(\min\left\{ \sqrt{\Phi(T)}, \Phi(T)/\sqrt{\eta} \right\}\right).
\end{equation}
\end{theorem}
\noindent Here, $\vol{(S)} \triangleq \sum_{v_i \in \mathcal{S}} degree(v_i)$ is the graph volume, and $\Phi(S)\triangleq \frac{\sum_{v_i\in \mathcal{S}} \sum_{v_j \in \mathcal{V} - \mathcal{S}} \mA(i,j)}{\min(\vol{(\mathcal{S})}, \vol{(\mathcal{V} - \mathcal{S})})}$ defines the conductance of a non-empty node set $\mathcal{S} \subset \mathcal{V}$.

\subsection{Finetuning Results for Supervised Learning}
\begin{table}[ht]
    \centering
    \caption{Fine-tuning results of mini-batch training with full labels. \textmd{We report accuracy(\%) for all datasets}.}
    \begin{tabular}{c|cccc}
    \toprule[1.1pt]
                & Arxiv & Products & MAG. & Papers100M  \\
    \midrule
    MLP & 55.50 & 61.06&39.11 & 47.24 \\
    SGC & 66.92 & 74.87 &54.68 & 63.29\\
    Random-Init$_{\small SAINT}$ &71.88 &81.15 &57.11 & 66.36\\
    Random-Init$_{\small LC}$ &71.47 &81.59 &56.84 & 66.47\\
    \midrule
    CCA-SSG & 72.24&80.61 &56.54 & 66.11\\
    GRACE &\underline{72.54} &\underline{82.45} &57.23 & 66.45\\
    BGRL &72.48& 82.37&57.34  & 66.57\\
      GraphMAE &72.38&81.89 &\underline{59.77}  &\textbf{66.64}  \\
    \midrule
      \model &\textbf{72.69} &\textbf{83.65} &\textbf{60.01} &\textbf{66.66} \\
    \bottomrule[1.1pt]
    \end{tabular}
    \label{tab:full}
\end{table}
Table~\ref{tab:full} shows the results of finetuning the pretrained encoder with all labels in the training set. It is observed that self-supervised learning can still lead to improvements in the supervised setting. 
\model can bring 1.22\%-3.17\% improvement compared to the randomly initialized model on three datasets, i.e., ogbn-Arxiv, Products, and MAG-Scholar-F. 
The only exception is ogbn-Papers100M, in which the improvement is only 0.2\% in accuray. The reason could be that the splitting strategy of this dataset (train:val:test = 78\%:8\%:14\%) resulted in an overly well-labeled training set. 

\subsection{Linear Probing Results on ogbn-Arxiv with Full-graph Training.}
\begin{table}[ht]
    \centering
    \caption{Linear probing results on ogbn-Arxiv with full-graph training. \textmd{We report accuracy(\%)}. }
    \begin{tabular}{c|c}
        \toprule[1.1pt]
                & Arxiv              \\
         \midrule
        GCN   & 71.74{\footnotesize $\pm$0.29}              \\
        GAT        & 72.10{\footnotesize $\pm$0.13}              \\
        \midrule
        DGI          & 70.34{\footnotesize $\pm$0.16}  \\
        GRACE       & 71.51{\footnotesize $\pm$0.11}  \\  
        BGRL       & 71.64{\footnotesize $\pm$0.12}        \\
        CCA-SSG   & 71.24{\footnotesize $\pm$0.20}    \\
        GraphMAE  & 71.75{\footnotesize $\pm$0.17}  \\
        \cmidrule{1-2}
        \model & {\bf 71.95}{\footnotesize $\pm$0.08} \\
        \bottomrule[1.1pt]
    \end{tabular}
    \label{tab:fullbatcharxiv}
\end{table}
As ogbn-Arxiv dataset is relatively small, we can conduct full batch training and inference. Table~\ref{tab:fullbatcharxiv} shows the results of \model as well as baselines under full batch training. Compared to GraphMAE, \model has a 0.2\% improvement. It is worth noting that GraphMAE results are better when trained with full-batch than with mini-batch, while the mini-batch training results of \model are comparable to the full-batch results.

\subsection{Implementation Notes}
\label{sec:impl}

\vpara{Running Environment.} 
Our proposed framework is implemented via PyTorch. For our methods, the experiments are conducted on a Linux machine with 1007G RAM, and 8 NVIDIA A100 with 80GB GPU memory. 
As for software versions, we use Python 3.9, PyTorch 1.12.0, OGB 1.3.3, and CUDA 11.3. 

\vpara{Model Configuration.}
For our model and all baselines, we pretrain the model using AdamW Optimizer with cosine learning rate decay without warmup. More details about pre-training hyper-parameters are in Table ~\ref{tab:hyper}.

\begin{table}[ht]
    \centering
    \caption{Hyper-parameters on large-scale  datasets.}
    \begin{tabular}{c|cccc}
    \toprule[1.1pt]
                & Arxiv & Products & MAG & Papers100M  \\
    \midrule
     masking rate & 0.5  & 0.5 & 0.5 &0.5 \\
     re-masking rate &  0.5 &0.5 & 0.5 & 0.5\\
     num\_re-masking & 3  &3 & 3 &3 \\
     coefficient $\lambda$ & 10.0  &5.0 & 0.1 & 10.0\\
     hidden\_size & 1024  &1024 &1024  & 1024\\
     num\_layer & 4  &4 & 4 & 4\\
     lr & 0.0025  &0.002 & 0.001 & 0.001\\
     weight\_decay & 0.06 & 0.06 & 0.04 &0.05 \\
     max epoch & 60  & 20& 10 &10 \\
    \bottomrule[1.1pt]
    \end{tabular}
    \label{tab:hyper}
    \vspace{-2mm}
\end{table}

\subsection{Baselines}
For large datasets, we choose four baselines GRACE~\cite{zhu2020deep}, BGRL~\cite{thakoor2021bootstrapped}, CCA-SSG~\cite{zhang2021canonical} and GraphMAE~\cite{hou2022graphmae}. To have a fair comparison, we download the public source code and use the GAT backbone. We adapted their code to integrate with sampling algorithms to run on large-scale graphs. For GGD~\cite{zheng2022rethinking}, we can only report the results available in the original paper since the authors have not released the code. The sources of the codes used are as follows:
\begin{itemize}
    \item BRGL: \url{https://github.com/Namkyeong/BGRL\_Pytorch}
    \item GRACE: \url{https://github.com/CRIPAC-DIG/GRACE}
    \item CCA-SSG: \url{https://github.com/hengruizhang98/CCA-SSG/}
    \item GraphMAE:\url{https://github.com/THUDM/GraphMAE}
\end{itemize}

\end{document}